\pdfoutput=1

\documentclass[11pt]{article}

\usepackage[preprint]{acl}

\usepackage{times}
\usepackage{latexsym}

\usepackage[T1]{fontenc}

\usepackage[utf8]{inputenc}

\usepackage{microtype}

\usepackage{inconsolata}

\usepackage{graphicx}
\usepackage{amsmath} 
\usepackage{hyperref}       
\usepackage{url}            
\usepackage{booktabs}       
\usepackage{amsfonts}       
\usepackage{nicefrac}       
\usepackage{microtype}      
\usepackage{xcolor}         

\usepackage{amssymb}
\usepackage{tocloft}
\usepackage{mathtools}
\usepackage{amsthm}
\usepackage{pifont}
\usepackage{newfloat}
\usepackage{listings}
\usepackage{algorithm}
\usepackage{algorithmic}
\usepackage[accsupp]{axessibility} 
\usepackage{adjustbox}
\usepackage{multirow}
\usepackage{booktabs}
\usepackage{tabularx}
\usepackage{color}
\usepackage{bm}
\usepackage{amsmath}
\usepackage{tikz}
\usepackage{xcolor}
\usepackage{tcolorbox}
\usepackage{wrapfig}
\title{FocusDiff: Advancing Fine-Grained Text-Image Alignment for Autoregressive Visual Generation through RL}

\author{
 Kaihang Pan$^{1\ast}$, Wendong Bu$^{1\ast}$, Yuruo Wu$^{1\ast}$, Yang Wu$^{2}$, Kai Shen$^{1}$, \\
\textbf{
Yunfei Li$^{2}$, Hang Zhao$^{2}$, Juncheng Li$^{1\dagger}$, Siliang Tang$^{1}$, Yueting Zhuang$^{1}$}  \\
$~\textsuperscript{\rm 1}$ Zhejiang University, 
$~\textsuperscript{\rm 2}$ Ant Group
\\\texttt {\{kaihangpan, wendongbu, shenkai, junchengli, siliang\}@zju.edu.cn}
\\ \texttt{wy306396@antgroup.com} \\
\url{https://focusdiff.github.io/}
}

\begin{document}
\maketitle

\renewcommand{\thefootnote}{\fnsymbol{footnote}}
\footnotetext[1]{\ \ Equal Contribution.}
\footnotetext[2]{\ \ Juncheng Li is the Corresponding Author.}
\renewcommand{\thefootnote}{\arabic{footnote}}

\begin{abstract}

Recent studies extend the autoregression paradigm to text-to-image generation, achieving performance comparable to diffusion models. However, our new PairComp benchmark -- featuring test cases of paired prompts with similar syntax but different fine-grained semantics -- reveals that existing models struggle with fine-grained text-image alignment thus failing to realize precise control over visual tokens. To address this, we propose FocusDiff, which enhances fine-grained text-image semantic alignment by focusing on subtle differences between similar text-image pairs. We construct a new dataset of paired texts and images with similar overall expressions but distinct local semantics, further introducing a novel reinforcement learning algorithm to emphasize such fine-grained semantic differences for desired image generation. Our approach achieves state-of-the-art performance on existing text-to-image benchmarks and significantly outperforms prior methods on PairComp.
\end{abstract}

\section{Introduction}

Witnessing the scalability of autoregression (AR) in large language models (LLMs~\citealp{openai2023chatgpt}), recent studies~\cite{sun2024autoregressive, chen2025janus} have extended the AR paradigm to text-to-image generation, achieving performance comparable to diffusion models. 
These methods employ visual tokenizers like VQGAN~\cite{esser2021taming} to discretize images, making them interpretable by LLMs as if they were a foreign language. 
After AR-based text-image alignment, image generation is transformed into a next-token-prediction task, harnessing the strong reasoning abilities of LLMs.

\begin{figure*}[t]
\includegraphics[width=\linewidth]{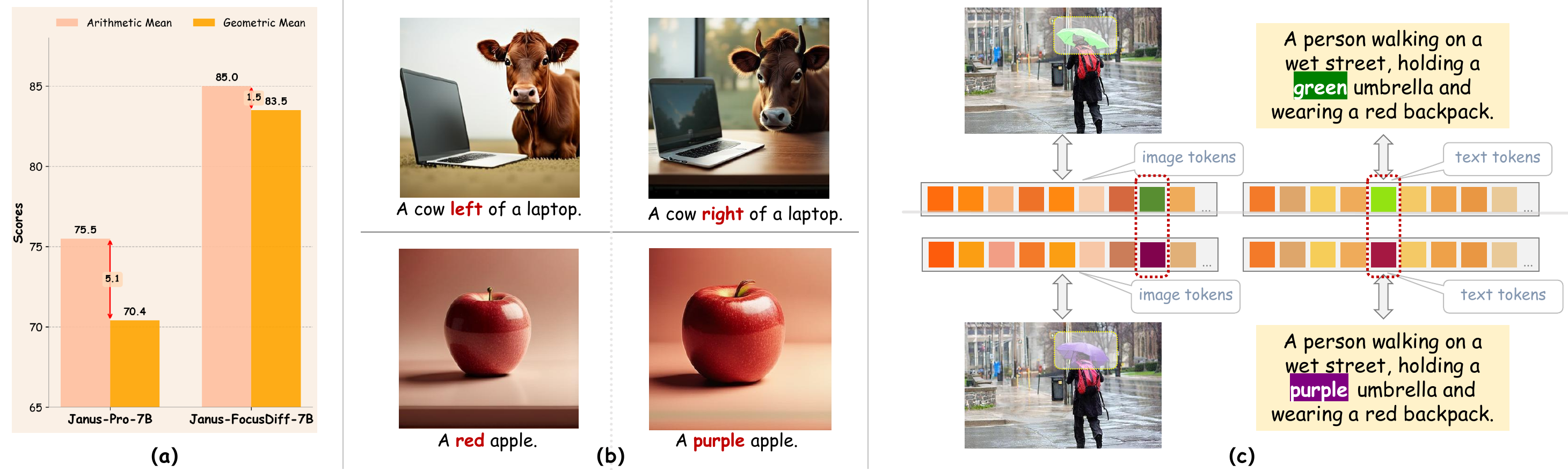}
\vspace{-2em}
\centering\caption{(a) For Janus-Pro-7B, the geometric mean score in PairComp is significantly lower than the arithmetic mean score (b) Examples of Janus-Pro-7B failing to generate images precisely according to the prompt. (c) The subtle sensory differences between images or between texts result in only minor alterations to specific tokens.}
\label{fig:intro}
\vspace{-1em}
\end{figure*}

Despite extensive vision-language alignment, existing models still struggle with precise control over visual tokens based on text conditions, leading to hallucination problems~\cite{vice2025exploring}. 
To further elucidate this problem, we first introduce the \textbf{PairComp benchmark}. 
Unlike typical text-to-image benchmarks~\cite{ghosh2023geneval} with a single prompt per test case, each case in PairComp consists of two prompts with similar syntactic but fine-grained semantic differences due to word-level distinctions.
For each prompt pair, we instruct text-to-image models to generate the image pairs and evaluate the text-image consistency scores $(s^1,s^2)$, with both the arithmetic and geometric means of $s^1$ and $s^2$ calculated.
Ideally, models should precisely distinguish the semantic nuances between prompts and accurately generate the corresponding images. 

However, even for the state-of-the-art AR model Janus-Pro~\cite{chen2025janus}, the geometric mean in PairComp is significantly lower than the arithmetic mean (Figure~\ref{fig:intro}.a).
Considering that the geometric mean is highly sensitive to lower values, the results indicate the instability of the AR model in fine-grained control over visual generation. 
The examples in Figure~\ref{fig:intro}.b further illustrate its inability to accurately control details such as object colors and spatial relationships. 
We argue that this problem lies in the lack of fine-grained text-image semantic alignment. 
Exhaustively covering all possible alignments for each text prompt is impractical, and images often contain irrelevant low-level semantics (\textit{e.g.}, background details that are not mentioned in the text)~\cite{ge2023making}. Thus, while current alignment ensures overall semantic coherence, it may introduce erroneous biases in fine-grained semantics, with some text tokens forming incorrectly alignment with several visual tokens.

Thus, a crucial question emerges:  \textit{\textbf{How can we achieve robust fine-grained text-image alignment to enable precise control over visual semantics in AR-based text-to-image generation?}}
Some studies~\cite{yin2024sea,zhao2024beyond} in multimodal comprehension leverage contrastive learning to build extra constraints for intra-sequence fine-grained token embedding alignment.
However, they undermine the core design philosophy of the decoder-only AR -- the causal dependency of tokens, failing to fully leverage the successful infrastructure of LLMs.
We aim to find an elegant solution for fine-grained text-image alignment without altering the original AR-based training paradigm.

We introduce \textbf{FocusDiff}, a method that enhances fine-grained text-image semantic alignment by learning from the differences between similar text-image pairs, without disrupting the original AR-based training paradigm.
Specifically, \textbf{from the data perspective}, we introduce \texttt{\textbf{FocusDiff-Data}}, expanding the training case from a single text-image pair $\{(\mathcal{T}, \mathcal{I})\}$ into a set of two pairs $\{(\mathcal{T}^1, \mathcal{I}^1,\mathcal{T}^2, \mathcal{I}^2 )\}$.
Here, $\mathcal{T}^1$ and $\mathcal{T}^2$, as well as $\mathcal{I}^1$ and $\mathcal{I}^2$,  appear similar in overall expression but differ in fine-grained details, with $\mathcal{T}^1$  being consistent with $\mathcal{I}^1$  but not with $\mathcal{I}^2$, and vice versa.
As shown in Figure~\ref{fig:intro}.c, the subtle sensory differences between images or between texts result in only minor alterations to specific visual or textual tokens.
Therefore, by comparing the token differences between these pairs, MLLM can trace how changes in text tokens lead to specific changes in visual tokens, establishing fine-grained semantic associations between the two modalities.

\textbf{From the training perspective}, we introduce Pair-GRPO, a reinforcement learning (RL) method that guides the model in learning fine-grained semantic differences through an exploration-exploitation trade-off. We formulate image generation as a Markov decision process and extend the GRPO framework~\cite{shao2024deepseekmath} to visual generation with a QA-based reward model, which eliminates the value function and estimates advantages in a group-relative manner.  We make two key improvements:

\textbf{(1) Expanding the Group Concept:} While vanilla GRPO considers $G$ responses from the same prompt as a group, we expand this to include $2\times G$ responses from pairs of similar prompts with fine-grained semantic differences from \texttt{FocusDiff-Data}.

\textbf{(2) Shifting Focus from Exploitation to Exploration:} 
Unlike vanilla GRPO, which encourages fully autonomous exploration without ground-truth images, we provide ground-truth images from \texttt{FocusDiff-Data} during early training to enhance exploration and guide the model to better grasp fine-grained semantic differences. As training progresses, we gradually reduce the reliance on these ground-truth images, transitioning from exploitation-first to exploration-first.

\begin{figure*}[t]
\includegraphics[width=\linewidth]{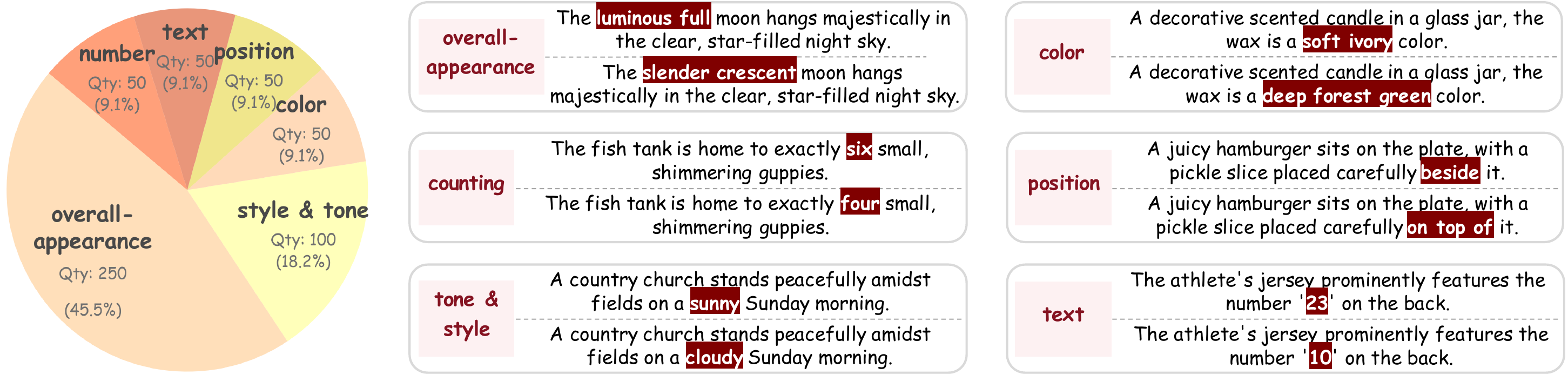}
\vspace{-1.5em}
\centering\caption{Statistical information of PairComp and test case examples for each subtask.}
\label{fig:paircomp}
\vspace{-1em}
\end{figure*}

Thanks to our novel training data and training strategy, with Janus-Pro as the backbone, we realize better fine-grained text-image semantic alignments and achieve precise control over visual semantics during text-to-image generation. Our main contributions are threefold:
\begin{itemize}
\item We introduce PairComp benchmark, featuring test cases with two prompts that share similar global expressions but differ in fine-grained semantics, which highlights existing models' limitations in precise visual control.
\item We propose FocusDiff, a paired text-image training dataset with an improved GRPO-based RL training paradigm, which focuses on fine-grained semantic differences to enhance text-image alignment.
\item We achieve SOTA performance on existing text-to-image benchmarks and significantly outperform prior methods on PairComp.
\end{itemize}

\section{Benchmark: PairComp}
\label{sec:paircomp}

\paragraph{Data format and Task Categorization.}
In traditional text-to-image benchmarks~\cite{ghosh2023geneval, huang2023t2i, hu2024ella}, each test case consists of a single prompt, which is used to measure the overall semantic alignment between the prompt and the image generated by the text-to-image model. 
In this section, we introduce a new benchmark called Paircomp. 
Each test case in Paircomp contains two similar prompts with subtle differences. 
By comparing the accuracy of the images generated by the model for each prompt, we evaluate whether the model has focused on the fine-grained semantic differences in the prompts to produce the corresponding correct images.
The two prompts in a test case exhibit word-level differences that lead to noticeable distinctions in certain fine-grained semantic aspects. These differences can be categorized into six types: (1) Overall appearance difference; (2) Color difference; (3) Counting difference; (4) Position difference; (5) Style \& Tone difference; (6) Text difference. In Figure~\ref{fig:paircomp}, we present examples for each category as well as statistical information. See more details in Appendix~\ref{app1}.

\paragraph{Evaluation Protocols.}
We use InternVL2.5-26B~\cite{chen2024internvl} as the evaluation model to assess the semantic consistency between the generated images and the text prompts. Specifically, for each image-prompt pair, we query the model with the prompt: ``Does this image match the description? Please directly respond with yes or no.'' We record the probability of the model responding with ``yes'' (denoted $P_{yes}$) and with ``no'' (denoted $P_{no}$), with the semantic consistency score calculated as $S(\mathcal{I}, \mathcal{T}) = P_{yes} / (P_{yes} + P_{no})$.

On this basis, given a subtask $\{(\mathcal{T}_i^1, \mathcal{T}_i^2)\}$, for each prompt pair, we instruct a text-to-image model to generate corresponding images $\{\mathcal{T}_i^1:(\mathcal{I}_i^{1,1},\mathcal{I}_i^{1,2}), \mathcal{T}_i^2:(\mathcal{I}_i^{2,1},\mathcal{I}_i^{2,2})\}_{i=1}^{N}$, with each prompt generating two images.
We define $s^{j,k}_i=S(\mathcal{I}_i^{j,k}, \mathcal{T}_i^{j})$, introduce two evaluation metrics: arithmetic mean 
$s_a = \frac{1}{4N} \sum_{i=1}^N\sum_{j=1}^2\sum_{k=1}^2 s_i^{j,k}$
, and geometric mean $s_g = \frac{1}{N} \sum_{i=1}^N \sqrt[4]{\prod_{j=1}^2\prod_{k=1}^2 s_i^{j,k}} $.
Here, the arithmetic mean measures the overall semantic alignment of the model with the prompts in the benchmark, while the geometric mean assesses the model's fine-grained precision and stability in generating images for similar prompts.

\section{Method: FocusDiff}

In this section, we introduce FocusDiff, a novel text-to-image method that focuses on the differences between similar text-image pairs to enhance fine-grained text-image alignment. 
From the data perspective, we propose \texttt{FocusDiff-Data}, expanding the training dataset from a single text-image pair to a set of two pairs. 
From the training perspective, we further propose Pair-GRPO, an improved RL framework that guides the model to focus on fine-grained semantic differences via an exploration-exploitation trade-off.

\subsection{Data Perspective: \texttt{FocusDiff-Data}}

Traditional text-to-image autoregressive training data comprises isolated text-image pairs lacking explicit connections. While ensuring global semantic alignment, it often fails to achieve fine-grained alignment. 
Images may contain redundant low-level information not mentioned in the text, and it is not practical to exhaustively cover all possible text-image alignments. 
Consequently, fine-grained alignment can be biased by confounders. 
For instance, if most apples in the training data are red, the model may incorrectly associate the color ``red'' with the word ``apple'', leading to a bias that hinders the generation of apples in other colors.

To address this issue, we turn to differential learning, which expands a single text-image pair $\{(\mathcal{T}, \mathcal{I})\}$ into two pairs $\{(\mathcal{T}^1, \mathcal{I}^1,\mathcal{T}^2, \mathcal{I}^2 )\}$. 
While $\mathcal{T}^1$ and $\mathcal{T}^2$, as well as $\mathcal{I}^1$  and $\mathcal{I}^1$, are similar in overall expression and global semantics, they differ in fine-grained details.
Consequently, $\mathcal{T}^1$ is semantically aligned with $\mathcal{I}^1$ but not with $\mathcal{I}^2$, and vice versa.
Given that both text and images are represented as discrete tokens in an AR framework, in fact, only a few token-level differences exist between $\mathcal{T}^1$ and $\mathcal{T}^2$, as well as between $\mathcal{I}^1$ and $\mathcal{I}^2$.
Then, the model is able to deduce how changes in text tokens lead to specific changes in visual tokens, allowing it to focus on subtle differences between texts and images, which ultimately enhances fine-grained text-image semantic alignment.

Then, the following question arises: How can we obtain such paired data, especially pairs of similar images? 
The image editing task, which involves before-and-after image pairs with localized changes, provides a feasible solution. 
We collect numerous before-and-after image pairs from \cite{yu2024anyedit} and \cite{zhao2024ultraedit}, covering a diverse range of editing types to reflect differences in various attributes. 
And then we can employ a powerful visual comprehension model like InternVL2.5-26B~\cite{chen2024internvl} to generate style-similar captions for the images.

Specifically, given the subpar quality of many image editing training datasets, we perform an initial screening to generate captions. Using the InternVL2.5-26B model, we assess three key aspects: (1) adherence to editing instructions, (2) consistency of non-edited areas with the original image, and (3) overall quality and natural appearance. We exclude any pairs that fail to meet these criteria.
Subsequently, we input the before-and-after-editing image pair and the editing instructions into InternVL2.5-26B, prompting it to generate captions with similar structure but differing key words to highlight the subtle image differences.

After generating the captions $(\mathcal{T}^1,\mathcal{T}^2)$ for the images $(\mathcal{I}^1,\mathcal{I}^2)$, we then perform a post-verification with three conditions:
(1) check if $\mathcal{T}^1$ and $\mathcal{T}^2$  have similar semantic structures; 
(2) verify the semantic alignment between $\mathcal{T}^1$ and $\mathcal{I}^1$, as well as between $\mathcal{T}^2$ and $\mathcal{I}^2$; 
 (3) ensure that $\mathcal{T}^1$ and $\mathcal{I}^2$, and $\mathcal{T}^2$ and $\mathcal{I}^1$ are not semantically aligned.
If all conditions are met, the sample is included in our training dataset. Otherwise, we use the InternVL2.5-26B model to regenerate captions and re-verify. If verification fails again, the image pair is discarded. Ultimately, we retained around $200,000$ high-quality data pairs for training to improve the model's capability to focus on fine-grained subtle differences. See more details in Appendix~\ref{app2}.

\begin{figure}[t]
\includegraphics[width=\linewidth]{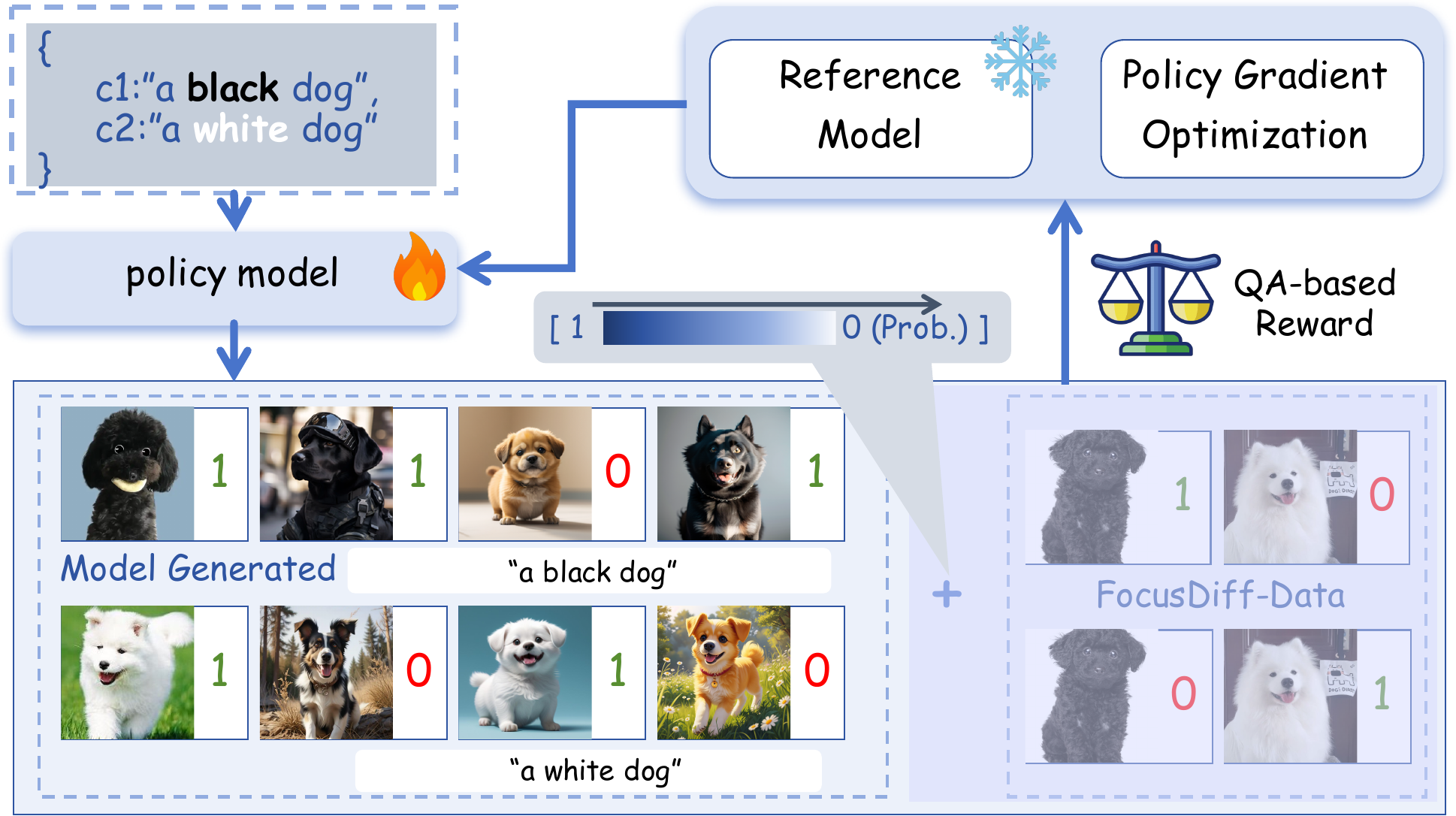}
\vspace{-1.5em}
\centering\caption{The framework of our Pair-GRPO.}
\label{fig:grpo}
\vspace{-1em}
\end{figure}

\subsection{Training Perspective: Pair-GRPO}

With \texttt{FocusDiff-Data}, we first conduct a supervised text-to-image fine-tuning. Subsequently, we treat image generation as a Markovian decision process at the token level and perform reinforcement learning based on an improved version of GRPO\cite{shao2024deepseekmath} (Figure~\ref{fig:grpo}), realizing a better balance of exploration-exploitation trade-off.

\paragraph{QA-based Reward.} 
The overall design philosophy of our reward model is to leverage a QA-based visual comprehension model (\textit{i.e.}, InternVL2.5-26B) to provide appropriate incentives, which will return a consistency score $\mathtt{R}_{\mathcal{I}}\in [0,1]$ for each text-image pair.
For example, for each prompt, we can generate questions for it via semantic decomposition, and ask the reward model to perform a VQA task given the prompt and the generated image, returning a score of 0 to 1 for each question. The reward is obtained by averaging the evaluations of the MLLMs on multiple questions for a prompt.

\begin{table*}[t]
    \centering
    \caption{ \label{tab:paircomp} Comparison with state-of-the-art models on our proposed PairComp. }
    \vspace{-1em}
    \resizebox{1.0\linewidth}{!}{
        \begin{tabular}{l|cc|cc|cc|cc|cc|cc|cc}
            \toprule
                    & \multicolumn{2}{c}{Overall Appear.}  &
                    \multicolumn{2}{c}{Color}  &
                    \multicolumn{2}{c}{Counting}  &
                    \multicolumn{2}{c}{Position}  &
                    \multicolumn{2}{c}{Style\&Tone}  &
                    \multicolumn{2}{c}{Text}  &
                    \multicolumn{2}{c}{\textbf{Average}}  \\
            Method  & $s_a\uparrow$ & $s_g\uparrow$ & $s_a\uparrow$ & $s_g\uparrow$ & $s_a\uparrow$ & $s_g\uparrow$ & $s_a\uparrow$ & $s_g\uparrow$ & $s_a\uparrow$ & $s_g\uparrow$ & $s_a\uparrow$ & $s_g\uparrow$ & $s_a\uparrow$ & $s_g\uparrow$\\
            \hline
            \multicolumn{15}{c}{\textit{Diffusion-based Method}} \\
            \hline

            PixArt-alpha~\cite{chen2023pixartalphafasttrainingdiffusion} & 75.9 &	68.7 &	88.8 &	86.2 &	62.8 &	58.4 &	54.2 &	49.3 &	87.8 &	84.5 &	35.4 &	28.8 &	67.5 &	62.7 \\ 
            SD3~\cite{esser2024scaling} & 82.5 &	77.0 &	95.4 &	94.6 &	74.0 &	70.3 &	71.9 &	68.5 &	89.4 &	86.2 &	\textbf{93.1} &	\textbf{92.0} &	84.4 &	81.4  \\
            FLUX.1-dev~\cite{flux2024} &78.7 &	71.1 &	94.3 &	92.0 &	63.9 &	60.0 &	70.4 &	66.1 &	84.4 &	79.7 &	90.1 &	85.4 &	80.3 &	75.7  \\
            Sana-1.5~\cite{xie2025sana} &83.8 &	79.5 &	97.3 &	96.8 &	\textbf{74.1} &	\textbf{71.5} &	69.1 &	64.0 &	92.7 &	90.1 &	82.4 &	77.9 &	83.2 &	80.0  \\
            Janus-Flow~\cite{ma2024janusflow} & 62.1 &	54.0 &	74.1 &	67.5 &	45.0 &	40.2 &	45.7 &	36.8 &	84.4 &	80.3 &	21.7 &	15.2 	&55.5 &	49.0 \\

            \hline
            \multicolumn{15}{c}{\textit{AR-based Method (including AR+diffusion)}} \\
            \hline
            
            LLamaGen~\cite{sun2024autoregressive}  & 53.5 &	45.4 &	67.0 &	61.2 &	45.3 &	39.5 &	42.1 &	35.4 &	68.8 &	60.1 &	18.0 &	12.0 &	49.1 &	42.3 \\
            VILA-U~\cite{wu2024vila} & 70.5 &	65.0 &	82.9 &	79.5 &	53.3 &	48.6 &	53.4 &	46.0 &	86.6 &	83.2 &	30.9 &	25.7 &	62.9 &	58.0 \\
            Show-o~\cite{xie2024show} & 68.5 &	62.2 &	87.2 &	85.0 &	58.2 &	55.2 &	45.2 &	40.6 &	87.8 &	84.7 &	34.9 &	26.8 &	63.6 &	59.1  \\
            SEED-X~\cite{ge2024seed} &83.2 	&79.7 &	95.5 	&94.5 &	64.9 &	62.3 &	63.0 	&59.9 &	90.0 &	87.3 	&52.2 &	45.1 &	74.8 &	71.5 \\
            Emu3~\cite{wang2024emu3} & 73.8 &	66.1 &	87.3 &	85.0 	&60.4 &	57.2 &	55.1 	&49.8 	&85.9 &	82.2 &	48.5 &	39.1 	&68.5 	&63.2\\ 
            VARGPTv1.1~\cite{zhuang2025vargpt} &59.5 &	51.6 &	80.5 &	77.6 &	37.2 &	32.6 &	42.2 &	35.2 &	82.5 &	79.1 &	19.4 &	13.6 &	53.6 &	48.3 \\
            Infinity~\cite{han2024infinity} &79.5 &	73.2 &	93.7 &	92.1 &	65.5 &	62.1 &	62.7 &	57.5 &	87.1 &	83.0 &	73.5 &	68.2 &	77.0 & 	72.7 \\
            BLIP3-o~\cite{chen2025blip3} &83.4 &	78.7 &	95.8 &	94.2 &	68.2 &	65.7 &	72.5 &	69.2 &	93.4 &	91.4 	&62.2 &	53.6 & 79.3 & 75.5\\
            Janus-Pro-1B~\cite{chen2025janus} & 75.6 &	69.5 &	89.7 &	87.7 &	36.1 	&29.5 	&56.2 &	50.2 	&92.3 &	90.4 &	37.6 	&28.0 &	64.6 	&59.2 \\
            Janus-Pro-7B~\cite{chen2025janus} & 82.3 	&75.6 &	95.7 &	94.0 &	52.7 	&47.1 &	69.4 &	63.9 	&92.0 &	88.7 &	60.8 &	53.2 &	75.5 &	70.4  \\
            T2I-R1~\cite{jiang2025t2i} & 84.6 &	80.3 &	96.5 	&95.9 	&68.1 &	65.2 	&71.3 	&67.5 	&91.2 	&89.2 &	82.5 &	77.5 &	82.4 &	79.3 \\
            Janus-Pro-R1~\cite{pan2025unlocking} &84.1 	&79.9 &	96.7 &	95.9 &	68.6 & 	65.8 &	71.9 &	70.0 &	93.3 &	91.8 &	77.1 &	71.6 &	82.0 &	79.2 \\

            \hline
            \textbf{Janus-FocusDiff-1B}  &78.4 	&75.0 &	91.7 &	90.0 &	49.5 &	44.4 &	64.7 &	61.8 &	91.9 &	90.8 &	49.8 &	46.3 &	71.0 &	68.1 \\
            \textbf{Janus-FocusDiff-7B}  &
            \textbf{85.4} &	\textbf{82.4} &	\textbf{97.8} &	\textbf{97.7} &	71.0 &	69.0 &	\textbf{75.9} &	\textbf{74.0} &	\textbf{94.3} &	\textbf{93.9} &	85.3 &	83.8 &	\textbf{85.0} &	\textbf{83.5}  \\

            \bottomrule
        \end{tabular}}
    \vspace{-1.2em}
\end{table*}

\paragraph{Vanilla GRPO for Autoregressive Image Generation.}
We adopt Group Relative Policy Optimization (GRPO) as the framework for reinforcement learning,
GRPO enhances PPO by eliminating the value function and estimating the advantages in a group-relative manner. 
Specifically, given the input prompt $\mathcal{T}$, the old policy $\pi_{\theta_{old}}$ first samples a group of $G$ individual images as the response group $\mathcal{G}=\{\mathcal{I}^1_i\}_{i=1}^G$.
We input each response with the group into the reward function to obtain the individual reward $\mathtt{R}_{\mathcal{I}_i}$.
We then calculate the advantages $\{A_i\}_{i=1}^G$, where each $A_i$ measures the relative quality of output compared to the average reward:
\begin{equation}
\small
\begin{aligned}
A_i = \frac{\mathtt{R}_{\mathcal{I}_i} - \text{mean}\big(\{ \mathtt{R}_{\mathcal{I}_i} \}_{i=1}^G\big)}{\text{std}\big(\{ \mathtt{R}_{\mathcal{I}_i} \}_{i=1}^G\big)}
\label{eq:adv}
\end{aligned}
\end{equation}

Then, we update the policy network parameters by the following training loss:
\begin{equation}
\footnotesize
\begin{aligned}
\mathcal{J}&(\theta) = \mathbb{E}_{\substack{(\mathcal{T},a)\sim \mathcal{D} \\ \{y_i\}_{i=1}^G\sim \pi_{\theta_\text{old}}(\cdot\mid \mathcal{T})}} 
\Bigg[ \frac{1}{\sum_{i=1}^{G}|y_i|}\sum_{i=1}^{G}\sum_{j=1}^{|y_i|} \Bigg(  \\
&\min \Big( \rho_{i,j} A_{i},
\text{clip} \Big( \rho_{i,j}, 1 - \varepsilon, 1 + \varepsilon \Big) A_{i} \Big) - \beta D_{\text{KL}}) 
\Bigg) \Bigg],
\label{eq:loss}
\end{aligned}
\end{equation}
where $D_{\text{KL}}=\frac{\pi_{ref}}{\pi}-\log \frac{\pi_{ref}}{\pi}-1$ is the the KL divergence to maintain training stability. And $\rho_{i,j}=\frac{\pi_{\theta}(y_{i,j} \mid \mathcal{T}, y_{i,<j})}{\pi_{\theta_{\text{old}}}(y_{i,j} \mid \mathcal{T},y_{i,<j})} $ is the ratio between probabilities of $\pi_\theta$ and $\pi_{\theta_{\text{old}}}$ for outputting current token.

\paragraph{Pair-GRPO for Fine-Grained Semantic Focusing.}

To enhance the model's ability to capture subtle differences between two prompts, we extend the group concept in GRPO from images generated by a single prompt to those generated by pairs of similar prompts. This aligns with our core idea of comparing the outputs of similar prompt pairs.
Specifically, give a pair of input prompt $\{\mathcal{T}^1, \mathcal{T}^2\}$ with similar global expressions but fine-grained semantics differences, a group of $G$ images $\{\mathcal{I}^1_i\}_{i=1}^G$ for $\mathcal{T}^1$ and another $G$ images $\{\mathcal{I}^2_i\}_{i=1}^G$ for $\mathcal{T}^2$ are sampled from the old policy.
And then $\{\mathcal{I}^1_i\}_{i=1}^G$ and $\{\mathcal{I}^1_2\}_{i=1}^G$ are assigned to the same group $\mathcal{G}_0 = \{(\mathcal{T}^1_i,\mathcal{I}^1_i)\}_{i=1}^G \cup \{(\mathcal{T}^2_i,\mathcal{I}^2_i)\}_{i=1}^G$ for advantage calculation.

Furthermore, from the $\texttt{FocusDiff-Data}$ dataset, we could also obtain the ground-truth images $\hat{\mathcal{I}}^1$ and $\hat{\mathcal{I}}^2$ corresponding to $\mathcal{T}^1$ and $\mathcal{T}^2$. 
Despite the high similarity between $\hat{\mathcal{I}}^1$ and $\hat{\mathcal{I}}^2$, during construction we ensure that $\hat{\mathcal{I}}^1$ achieves a favorable reward score when conditioned on $\mathcal{T}^1$, but achieves an unfavorable score when conditioned on $\mathcal{T}^2$. 
Thus, if we further incorporate $\hat{\mathcal{I}}^1$ into the group, it can assume a dual role within the group: it serves as a positive guide in $\{(\mathcal{T}^1_i,\mathcal{I}^1_i)\}_{i=1}^G$ indicating to the model about the correct visual semantics, and as a cautionary counterexample in $\{(\mathcal{T}^2_i,\mathcal{I}^2_i)\}_{i=1}^G$, warning the model to avoid generating erroneous visual semantics that are commonly encountered.
The same applies to $\hat{\mathcal{I}}^2$.

On this basis, we introduce a dynamic probability $p$ that starts at $1.0$ and gradually decreases to $0.0$ during RL training. 
At each training iteration, with probability $p$, we expand the group $\mathcal{G}$ to include the above additional pairs from \texttt{FocusDiff-Data}:
$\mathcal{G} = \mathcal{G}_0 + \{(\mathcal{T}^1,\hat{\mathcal{I}}^1),(\mathcal{T}^1,\hat{\mathcal{I}}^2),(\mathcal{T}^2,\hat{\mathcal{I}}^1), (\mathcal{T}^2,\hat{\mathcal{I}}^2)\}$. 
Otherwise, the group remains as $\mathcal{G} = \mathcal{G}_0$. 
\textbf{This is a process of shifting focus from exploitation to exploration.} 
In the early stages of training, the labeled images from the dataset encourage more exploitation to the model, offering more appropriate guidance. 
As training progresses and the model's ability to grasp fine-grained differences strengthens, the probability of providing labeled images gradually decreases. 
We simply provide the model with the right incentives, encouraging it to develop advanced problem-solving strategies through fully autonomous exploration.

In each iteration, after defining the group concept, we employ the same way as Eq.(\ref{eq:adv}) to calculate the advantages. Finally, the objective function is consistent with Eq.(\ref{eq:loss}).

\begin{table*}[t]
    \centering
    \caption{ \label{tab:t2i} Comparison with state-of-the-art models on GenEval, T2I-CompBench and DPG-Bench on zero-shot text-to-image generation. The best results are in \textbf{bold fonts} with the second best \underline{underlined}. }
    \vspace{-1em}
    \resizebox{1.0\linewidth}{!}{
        \begin{tabular}{l|ccccccc|ccc|c}
            \toprule
                    & \multicolumn{7}{c}{\textbf{GenEval}}  &  \multicolumn{3}{c}{\textbf{T2I-CompBench}} &  \multicolumn{1}{c}{\textbf{DPG-Bench}}   \\
            Method  & \textbf{Overall}$\uparrow$  & SingObj$\uparrow$ &  TwoObj$\uparrow$ & Counting$\uparrow$  & Color$\uparrow$ & Pos. $\uparrow$ & ColorAttr $\uparrow$ &  Color$\uparrow$  & Shape$\uparrow$ & Texture$\uparrow$ & Avg$\uparrow$\\
            \hline
            \multicolumn{12}{c}{\textit{Diffusion-based Method}} \\
            \hline
            PixArt-alpha~\cite{chen2023pixartalphafasttrainingdiffusion}       & 0.48 & 0.98 & 0.50 & 0.44 & 0.80 & 0.08 & 0.07    & 68.9  & 55.8 & 70.4   & 71.11\\
            DALL-E 3 ~\cite{betker2023improving}   & 0.67  & 0.96 & 0.87 & 0.47 & 0.83 & 0.43 & 0.45    & 81.1 & \textbf{67.5} & \textbf{80.7}  & 83.50\\ 
            SD3~\cite{esser2024scaling} &  0.74      & 0.99 & 0.94 & 0.72 & 0.89 & 0.33 & 0.60  & - & - & - & 84.08\\
            FLUX.1-dev~\cite{flux2024} & 0.66 & 0.98 & 0.79 & 0.73 & 0.77 & 0.22 & 0.45 & - & - & -  & 83.79 \\
            Sana-1.5~\cite{xie2025sana} & 0.81 & 0.99 & 0.93 & \textbf{0.86} & 0.84 & 0.59 & 0.65 & - & - & -  & 84.70 \\
            Janus-Flow~\cite{ma2024janusflow} & 0.63 & 0.97 & 0.59 & 0.45 & 0.83 & 0.53 & 0.42 & - & - & -  & 80.09 \\

            \hline
            \multicolumn{12}{c}{\textit{AR-based method}} \\
            \hline
            LLaMAGen~\cite{sun2024autoregressive} & 0.32 & 0.71 & 0.34 & 0.21 & 0.58 & 0.07 & 0.04 & - & - & -  & 65.16 \\
            VILA-U~\cite{wu2024vila} & 0.40 & 0.88 & 0.42 & 0.25 & 0.69 & 0.08 & 0.09 & 56.8 & 43.3 & 50.1  & -\\
            Show-o~\cite{xie2024show} & 0.68 & 0.98 & 0.80 & 0.66 & 0.84 & 0.31 & 0.50 & 56.0 & 41.0 & 46.0  & 67.48 \\
            SEED-X~\cite{ge2024seed} &  0.49 &0.96 & 0.57 & 0.29 & 0.82 & 0.14 & 0.15  & 65.7 & 49.2 & 60.3  & - \\
            Emu3~\cite{wang2024emu3}  & 0.54 & 0.98 & 0.71 & 0.34 & 0.81 & 0.17 & 0.21  & 61.1 & 47.3 & 61.8  & 80.60\\ 
            DDT-LLaMA~\cite{pan2025generative} & 0.66 & 0.99 & 0.64 & 0.56 & 0.87 & 0.39 & 0.48 & 72.8 & 51.4 & 64.2  & 80.90 \\
            VARGPTv1.1~\cite{zhuang2025vargpt} & 0.53 & 0.96 & 0.53 & 0.48 & 0.83 & 0.13 & 0.21 & - & - & -  & 78.59 \\
            Infinity~\cite{han2024infinity} & 0.73 & - & 0.85 & - & - & 0.49 & 0.57 & - & - & -  & 83.46 \\
            BLIP3-o-8B~\cite{chen2025blip3} & 0.84 & - & - & - & - & - & - & 79.7 & 52.8 & 68.0  & 81.60 \\
            GPT-4o~\cite{gpt40} & 0.85 & 0.99 & 0.92 & 0.85 & 0.91 & 0.75 & 0.66 & - & - & -  & - \\
            Janus-Pro-1B~\cite{chen2025janus} & 0.73 & 0.98 & 0.82 & 0.51 & 0.89 & 0.65 & 0.56 & 55.1 & 37.8 & 47.6  & 82.63 \\
            Janus-Pro-7B~\cite{chen2025janus} & 0.80 & 0.99 & 0.89 & 0.59 & 0.90 & 0.79 & 0.66 & 63.6 & 35.3 & 49.4  & 84.17 \\

            \hline
            \multicolumn{12}{c}{\textit{AR-based Method + RL}} \\
            \hline
            Show-o+PARM~\cite{guo2025generateimagescotlets} & 0.69 & 0.97 & 0.75 & 0.60 & 0.83 & 0.54 & 0.53 & 75.0 & 56.0 & 66.0  & - \\
            T2I-R1~\cite{jiang2025t2i} & 0.79 & 0.99 & 0.91 & 0.53 & 0.91 & 0.76 & 0.65 & 81.3 & 58.5 & 72.4  & 84.42 \\
            \hline
            \textbf{Janus-FocusDiff-1B}  & 0.82 & 0.99 & 0.93 & 0.59 & 0.90 & 0.80 & 0.68 & 61.5 & 47.7 & 60.4  & 83.17 \\
            \textbf{Janus-FocusDiff-7B}  & \textbf{0.85} & \textbf{0.99} & \textbf{0.95} & 0.63 & \textbf{0.93} & \textbf{0.85} & \textbf{0.75} & \textbf{83.0} & \underline{60.3} & \underline{72.8}  & \textbf{85.23}\\

            \bottomrule
        \end{tabular}
    }
    \vspace{-1.2em}
\end{table*}

\section{Experiments}

We employ Janus-Pro~\cite{chen2025janus} as the backbone, developing Janus-FocusDiff, excelling in text-to-image generation, with improved capabilities of vision-language alignment. More details are given in Appendix \ref{app3} and \ref{app4}.

\subsection{Main Results on PairComp}
We first conduct zero-shot evaluations on PairComp for our model and recent advanced diffusion-based and AR-based text-to-image methods (including those integrating AR with diffusion).
Following the evaluation protocols in \S~\ref{sec:paircomp}, we report the arithmetic mean scores $s_a$ and geometric mean scores $s_g$ of these methods in Table~\ref{tab:paircomp}.
First, we have the following key findings of existing methods:

\textbf{(1) The overall text-image alignment is satisfactory.} 
Existing leading models, both AR-based and diffusion-based, exhibit relatively high arithmetic mean scores.
And the diffusion-based SOTA models, SD3~\cite{esser2024scaling} and Sana-1.5~\cite{xie2025sana}, achieve higher average performance than the AR-based SOTA models, T2I-R1~\cite{jiang2025t2i} and Janus-Pro-R1~\cite{pan2025unlocking}.

\textbf{(2) The stability of image generation is poor,}
making it difficult to precisely control over fine-grained visual semantics that reflect subtle differences specified in the prompts. 
The gap between the geometric mean and the arithmetic mean reflects the stability of a model's image generation. 
It can be seen that current methods struggle to obtain ideal geometric mean scores. 
The average $s_g$ of SD3 is 3.0 points lower than its $s_a$, and the average $s_g$ of Janus-Pro-7B is 5.1 points lower than its $s_a$
This indicates poor stability in image generation without precise control over visual semantics.
Meanwhile, It is also worth noting that AR-based methods exhibit slightly lower stability in image quality compared to diffusion-based methods.

Compared to existing methods, Janus-FocusDiff-7B achieves the following advantages:
\textbf{(1) Improved text-image alignment} is achieved with higher arithmetic mean scores. 
After training, we enhance Janus-Pro-7B to achieve better global vision-language alignment, with the average performance in PairComp surpassing that of the previous SOTA, SD3 ($85.0$ vs. $84.4$ in $s_a$, $83.5$ vs. $81.4$ in $s_g$).
Compared to the backbone model Janus-Pro-7B, the average values of $s_a$ and $s_g$ have achieved substantial improvements of 9.5 and 13.1 points, respectively.
Furthermore, when compared to T2I-R1 and Janus-Pro-R1, baseline models that similarly employ reinforcement learning based on Janus-Pro-7B, Janus-FocusDiff-7B also demonstrates superior performance across all sub-tasks.
\textbf{(2) Enhanced Generation Stability} is achieved with a significantly reduced gap between $s_a$ and $s_g$, with only an average 1.5-point difference, hich is far smaller than the gap between $s_a$ and $s_g$  observed in other baseline models. This further demonstrates that our method achieves better fine-grained text-image semantic alignment, allowing the MLLM to focus on the subtle semantic differences in prompts for stable, high-quality image generation.

\begin{figure*}[t]
\includegraphics[width=\linewidth]{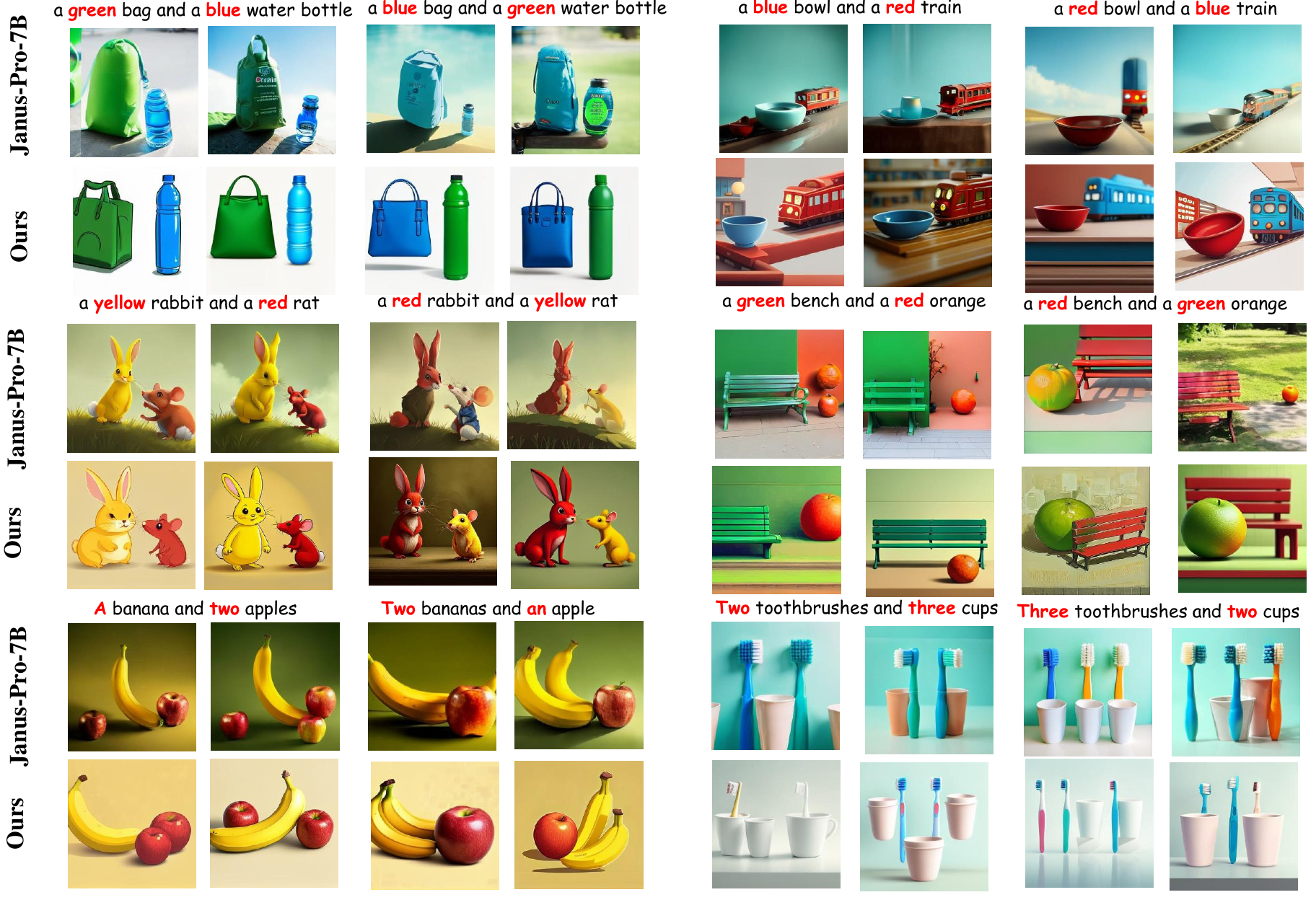}
\vspace{-2em}
\centering\caption{Qualitative Comparisons between Janus-Pro-7B and our Janus-FocusDiff on pairs of similar prompts. For each prompt, we ask each model to generate two images.}
\label{fig:compare1}
\vspace{-1em}
\end{figure*}

\subsection{Main Results on Existing Benchmarks}
And then we further conduct zero-shot evaluation on 3 text-to-image benchmarks: GenEval~\cite{ghosh2023geneval}, T2I-CompBench~\cite{huang2023t2i}, and DPG-Bench~\cite{hu2024ella}. 
The comparison results against both diffusion-based and MLLM-based methods are presented in Table~\ref{tab:t2i}.
We have the following observations:

\textbf{(1)} In most settings, our model outperforms other diffusion-based and MLLM-based baselines, achieving SOTA performance. For example, on the GenEval benchmark, the overall performance of Janus-Pro-R1 is even on par with that of GPT-4o. This underscores that we endow the MLLM with enhanced capability of vision-language alignment.
\textbf{(2)} Compared to other baselines that also propose incorporating RL into AR-based text-to-image generation, our method achieves superior performance. For example, it consistently outperforms the concurrent work T2I-R1 on the T2i-Compbench with the same backbone model. This highlights the effectiveness of our pair-GRPO algorithm.
\textbf{(3)} Compared to the backbone model Janus-Pro-7B, our method achieves performance improvements of 6.3\% on Geneval, 45.6723\% on T2i-Compbench, and 1.3\% on DPG-bench, respectively. These results underscore the effectiveness of our approach, which significantly enhances the text-to-image generation capabilities of the base model.

\subsection{Qualitative Comparisons}

\begin{figure*}[t]
\includegraphics[width=\linewidth]{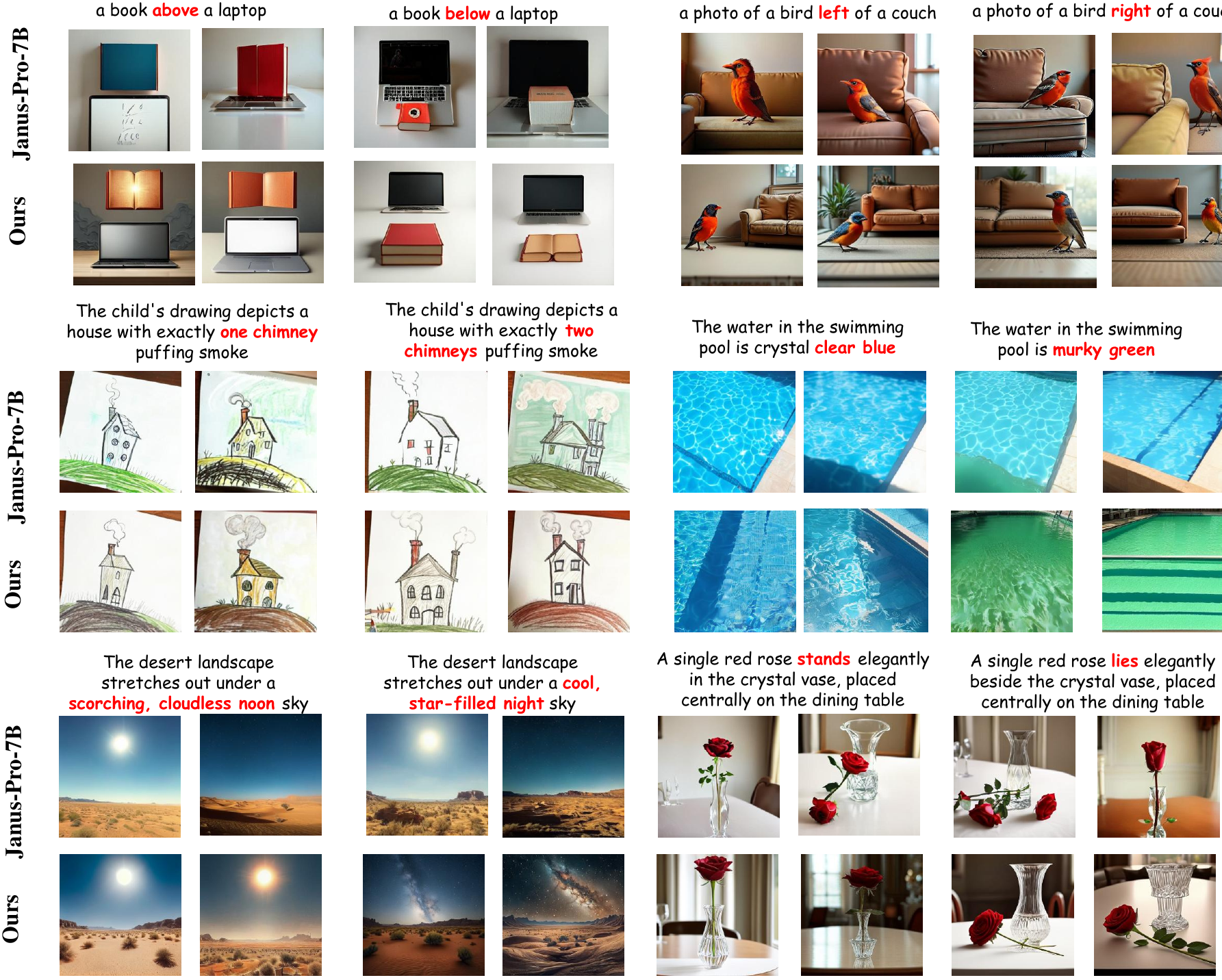}
\vspace{-1.5em}
\centering\caption{More qualitative Comparisons between Janus-Pro-7B and Janus-FocusDiff on pairs of similar prompts.}
\label{fig:compare2}
\vspace{-1em}
\end{figure*}

Figures~\ref{fig:compare1} and~\ref{fig:compare2}  present a direct qualitative comparison between Janus-FocusDiff-7B and Janus-Pro-7B on pairs of similar prompts with fine-grained semantic differences.
For each prompt, we ask each model to generate two images. 
We can see that Janus-Pro-7B struggles to precisely control the fine-grained requirements of similar prompts. 
Moreover, even for the same prompt, the generated images are not consistently aligned with the target semantics. 
In contrast, our Janus-FocusDiff-7B is capable of accurately capturing the fine-grained semantic differences between prompts to generate corresponding images and stably produces high-quality images that meet the specified requirements.

\begin{figure*}[t]
\includegraphics[width=\linewidth]{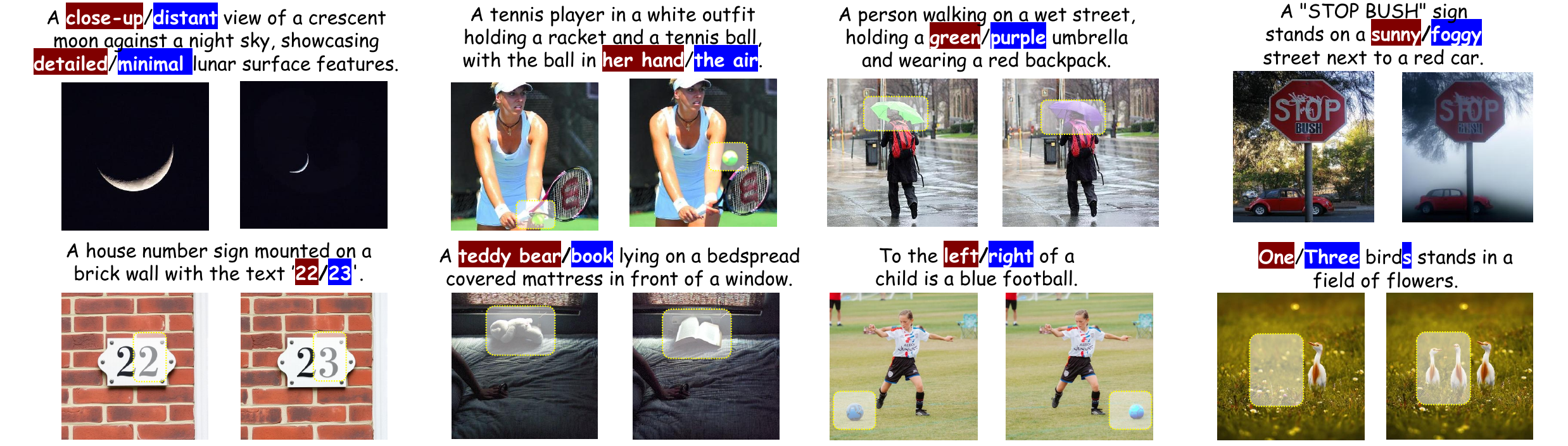}
\vspace{-1.5em}
\centering\caption{Examples of training data in \texttt{FocusDiff-Data}.}
\label{fig:focusdiff-data-case}
\vspace{-1em}
\end{figure*}

\begin{figure}[t]
\includegraphics[width=\linewidth]{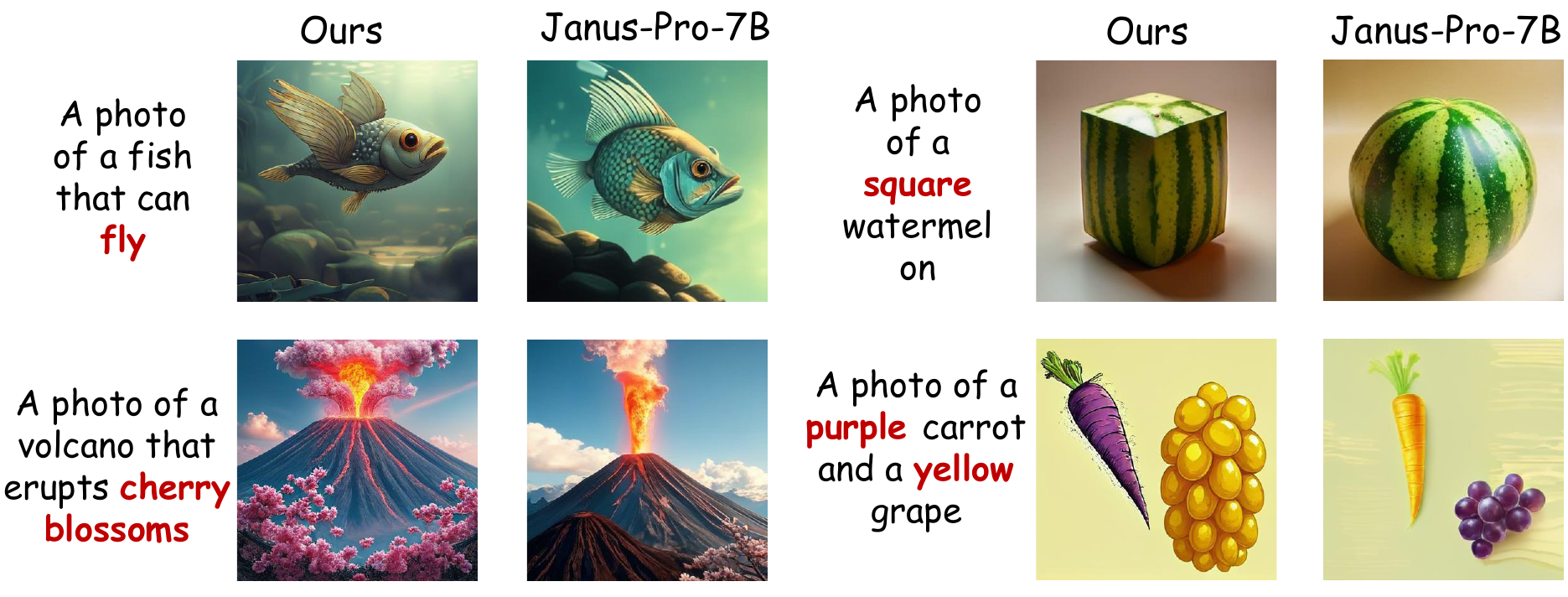}
\vspace{-2em}
\centering\caption{Counterfactual image generation.}
\label{fig:cf}
\vspace{-1em}
\end{figure}

\subsection{In-depth Analysis}

\paragraph{Effect of Pair-GRPO.} 
To demonstrate the superiority of the Pair-GRPO algorithm, we trained the following ablation models:
\textbf{(1) w/o Group Expanding:} The group concept is restricted to images generated from a single prompt.
\textbf{(2) w/o GT Image:} We set $p=0.0$ and do not provide ground-truth images during RL.
\textbf{(3) Vanilla GRPO:} We fully degrade Pair-GRPO to the vanilla GRPO.
As shown in Table~\ref{tab:abla} \textbf{Rows3-5}, Pair-GRPO consistently outperforms the other ablated algorithms on both Geneval and Paircomp. This indicates that Pair-GRPO is more effective in focusing on the fine-grained prompt requirements, thereby generating images that better align with the intended prompt semantics.

\begin{table}[t]
    \centering
    \caption{ \label{tab:abla} Ablation Study on GenEval and PairComp. }
    
    \vspace{-1em}
    \resizebox{1.0\linewidth}{!}{
        \begin{tabular}{ll|c|cccc}
            \toprule
                &    & GenEval  &  \multicolumn{2}{c}{PairComp-Overall} &  \multicolumn{2}{c}{PairComp-Avg}   \\
       &   Methods & \textbf{Overall}$\uparrow$ & $s_a\uparrow$ & $s_g\uparrow$& $s_a\uparrow$& $s_g\uparrow$ \\ \hline
1 & Janus-Pro-7B & 0.80& 82.3 & 75.6 & 75.5 & 70.4\\
2& Janus-FocusDiff-7B & \textbf{0.85}& \textbf{85.4} & \textbf{82.4} & \textbf{85.0} & \textbf{83.5}\\ \hline
   3& w/o Group Expanding & 0.84& 84.6 & 79.8 & 83.0& 79.8\\
   4& w/o GT Image & 0.84&  84.9 & 81.3 & 84.1 & 82.0\\
   5& Vanilla GRPO & 0.83&  83.6 & 77.6 &80.7 & 76.6\\ \hline
   6& w/o \texttt{FocusDiff-Data} & 0.82& 82.7 & 76.0 & 77.0 & 71.9\\
            \bottomrule
        \end{tabular}
    }
    \vspace{-1.2em}
\end{table}

\paragraph{Effect of \texttt{FocusDiff-Data}.}
We further generate a set of prompts commonly used in text-to-image to replace \texttt{FocusDiff-Data} for RL  training with Vanilla GRPO. 
As shown in Table~\ref{tab:abla} \textbf{Rows5-6}, with GRPO as the RL framework, the performance obtained from training with \texttt{Focusdiff-Data} outperforms that derived from training with the newly generated prompts. 
This indicates that \texttt{FocusDiff-Data} enables the model to achieve better text-image alignment by focusing on the subtle semantic differences between similar prompts.

\paragraph{Effect of Model Scale.}
Given that Janus-Pro-7B already possesses formidable image generation capabilities, to further investigate the effectiveness of FocusDiff, we employ Janus-Pro-1B as the backbone and conduct training under the same settings to develop Janus-FocusDiff-1B.
As shown in Tables~\ref{tab:t2i} and \ref{tab:paircomp},
Janus-FocusDiff-1B demonstrates significant performance improvements compared to Janus-Pro-1B across all of four benchmarks (\textit{e.g.}, 12.3\% on Geneval, 20.7\% on T2i-CompBench, 12.4\% on PairComp). 
It even outperforms Janus-Pro-7B on GenEval and T2I-CompBench, further validating the effectiveness of our approach.

\paragraph{Examples of \texttt{FocusDiff-Data}.}
In Figure~\ref{fig:focusdiff-data-case}, we present some  cases in \texttt{FocusDiff-Data} to intuitively demonstrate the dataset's advantages. It is evident that the images and their corresponding prompts exhibit only region-level or word-level differences. This design enables models to focus on learning fine-grained semantic alignment between text and images.

\paragraph{Image Generation with Counterfactual Prompts}

Endowing the model with fine-grained control over visual details, it can further generate images that more accurately match counterfactual prompts that are rarely found in the real world, as shown in Figure~\ref{fig:cf}. For example, given the prompt   ``square watermelon'', Janus-Pro-7B still generates a round one. 
In contrast, our Janus-FocusDiff successfully generates a watermelon with this counterfactual shape. 
This indicates that we effectively mitigate the issue of hallucination generation, eliminating the erroneous bias toward the training distribution.

\section{Related Work}

In recent years, diffusion models~\cite{flux2024, esser2024scaling} have dominated the realm of visual generation. However, recent efforts have explored using autoregressive (AR) models~\cite{wang2024emu3, sun2024autoregressive, chen2025janus, pan2024auto, han2024infinity} to generate images by predicting the next token in a sequence and have achieved comparable performance. 
These methods typically tokenize images into discrete codes using VQ-VAE~\cite{esser2021taming}.
Subsequently, a decoder-only transformer is trained for text-image alignment, predicting image codes that are then detokenized back into images.
Furthermore, the AR property satisfies the optimality condition of policy improvement, which further supports effective post-training based on RL for visual generation~\cite{guo2025generateimagescotlets, jiang2025t2i, lin2025reasoning, pan2025unlocking}, similar to LLM~\cite{guo2025deepseek}.
However, most existing methods focus primarily on the overall semantics, struggling with fine-grained text-image alignment. 
In contrast, our FocusDiff enables AR-based models to achieve precise control over visual tokens for stable and high-quality image generation.

\section{Conclusion}

In this paper, we propose \textbf{PairComp}, a new benchmark for text-to-image generation revealing that existing models struggle with fine-grained text-image alignment. And we introduce \textbf{FocusDiff}, a training paradigm with a novel training dataset and an improved RL algorithm, enhancing fine-grained text-image semantic alignment by focusing on subtle differences between similar text-image pairs.
On this basis, we develop Janus-FocusDiff, achieving SOTA performance on existing text-to-image benchmarks and significantly outperforms prior methods on PairComp.

\section*{Limitations}
In the \texttt{FocusDiff-Data}, there is a limited coverage of prompts related to ``counting'' or ``text'', which has led to relatively lower performance of Janus-FocusDiff in the  ``Counting'' and  ``Text'' subtasks of Paircomp, as well as the   ``counting'' subtask of Geneval. To address this issue, in the future, we plan to expand the scale of \texttt{FocusDiff-Data} to include more prompts and images related to ``counting'' or ``text'', and to further diversify the types of prompts covered.


\bibliography{acl_latex}

\begin{thebibliography}{48}
\providecommand{\natexlab}[1]{#1}

\bibitem[{Bai et~al.(2023)Bai, Bai, Chu, Cui, Dang, Deng, Fan, Ge, Han, Huang, Hui, Ji, Li, Lin, Lin, Liu, Liu, Lu, Lu, Ma, Men, Ren, Ren, Tan, Tan, Tu, Wang, Wang, Wang, Wu, Xu, Xu, Yang, Yang, Yang, Yang, Yao, Yu, Yuan, Yuan, Zhang, Zhang, Zhang, Zhang, Zhou, Zhou, Zhou, and Zhu}]{qwen}
Jinze Bai, Shuai Bai, Yunfei Chu, Zeyu Cui, Kai Dang, Xiaodong Deng, Yang Fan, Wenbin Ge, Yu~Han, Fei Huang, Binyuan Hui, Luo Ji, Mei Li, Junyang Lin, Runji Lin, Dayiheng Liu, Gao Liu, Chengqiang Lu, Keming Lu, and 29 others. 2023.
\newblock Qwen technical report.
\newblock \emph{arXiv preprint arXiv:2309.16609}.

\bibitem[{Bengio et~al.(2009)Bengio, Louradour, Collobert, and Weston}]{bengio2009curriculum}
Yoshua Bengio, J{\'e}r{\^o}me Louradour, Ronan Collobert, and Jason Weston. 2009.
\newblock Curriculum learning.
\newblock In \emph{Proceedings of the 26th annual international conference on machine learning}, pages 41--48.

\bibitem[{Betker et~al.(2023)Betker, Goh, Jing, Brooks, Wang, Li, Ouyang, Zhuang, Lee, Guo et~al.}]{betker2023improving}
James Betker, Gabriel Goh, Li~Jing, Tim Brooks, Jianfeng Wang, Linjie Li, Long Ouyang, Juntang Zhuang, Joyce Lee, Yufei Guo, and 1 others. 2023.
\newblock Improving image generation with better captions.
\newblock \emph{Computer Science. https://cdn. openai. com/papers/dall-e-3. pdf}, 2(3):8.

\bibitem[{Chen et~al.(2024{\natexlab{a}})Chen, Pan, Dai, Wang, Zhuang, Tang, and Xu}]{chen2024improving}
Dong Chen, Kaihang Pan, Guangyu Dai, Guoming Wang, Yueting Zhuang, Siliang Tang, and Mingliang Xu. 2024{\natexlab{a}}.
\newblock Improving vision anomaly detection with the guidance of language modality.
\newblock \emph{IEEE Transactions on Multimedia}.

\bibitem[{Chen et~al.(2025{\natexlab{a}})Chen, Xu, Pan, Hu, Qin, Goldstein, Huang, Zhou, Xie, Savarese et~al.}]{chen2025blip3}
Jiuhai Chen, Zhiyang Xu, Xichen Pan, Yushi Hu, Can Qin, Tom Goldstein, Lifu Huang, Tianyi Zhou, Saining Xie, Silvio Savarese, and 1 others. 2025{\natexlab{a}}.
\newblock Blip3-o: A family of fully open unified multimodal models-architecture, training and dataset.
\newblock \emph{arXiv preprint arXiv:2505.09568}.

\bibitem[{Chen et~al.(2023)Chen, Yu, Ge, Yao, Xie, Wu, Wang, Kwok, Luo, Lu, and Li}]{chen2023pixartalphafasttrainingdiffusion}
Junsong Chen, Jincheng Yu, Chongjian Ge, Lewei Yao, Enze Xie, Yue Wu, Zhongdao Wang, James Kwok, Ping Luo, Huchuan Lu, and Zhenguo Li. 2023.
\newblock \href {https://arxiv.org/abs/2310.00426} {Pixart-$\alpha$: Fast training of diffusion transformer for photorealistic text-to-image synthesis}.
\newblock \emph{Preprint}, arXiv:2310.00426.

\bibitem[{Chen et~al.(2025{\natexlab{b}})Chen, Wu, Liu, Pan, Liu, Xie, Yu, and Ruan}]{chen2025janus}
Xiaokang Chen, Zhiyu Wu, Xingchao Liu, Zizheng Pan, Wen Liu, Zhenda Xie, Xingkai Yu, and Chong Ruan. 2025{\natexlab{b}}.
\newblock Janus-pro: Unified multimodal understanding and generation with data and model scaling.
\newblock \emph{arXiv preprint arXiv:2501.17811}.

\bibitem[{Chen et~al.(2024{\natexlab{b}})Chen, Wu, Wang, Su, Chen, Xing, Zhong, Zhang, Zhu, Lu et~al.}]{chen2024internvl}
Zhe Chen, Jiannan Wu, Wenhai Wang, Weijie Su, Guo Chen, Sen Xing, Muyan Zhong, Qinglong Zhang, Xizhou Zhu, Lewei Lu, and 1 others. 2024{\natexlab{b}}.
\newblock Internvl: Scaling up vision foundation models and aligning for generic visual-linguistic tasks.
\newblock In \emph{Proceedings of the IEEE/CVF Conference on Computer Vision and Pattern Recognition}, pages 24185--24198.

\bibitem[{Esser et~al.(2024)Esser, Kulal, Blattmann, Entezari, M{\"u}ller, Saini, Levi, Lorenz, Sauer, Boesel et~al.}]{esser2024scaling}
Patrick Esser, Sumith Kulal, Andreas Blattmann, Rahim Entezari, Jonas M{\"u}ller, Harry Saini, Yam Levi, Dominik Lorenz, Axel Sauer, Frederic Boesel, and 1 others. 2024.
\newblock Scaling rectified flow transformers for high-resolution image synthesis.
\newblock In \emph{Forty-first international conference on machine learning}.

\bibitem[{Esser et~al.(2021)Esser, Rombach, and Ommer}]{esser2021taming}
Patrick Esser, Robin Rombach, and Bjorn Ommer. 2021.
\newblock Taming transformers for high-resolution image synthesis.
\newblock In \emph{Proceedings of the IEEE/CVF conference on computer vision and pattern recognition}, pages 12873--12883.

\bibitem[{Ge et~al.(2023)Ge, Zhao, Zeng, Ge, Li, Wang, and Shan}]{ge2023making}
Yuying Ge, Sijie Zhao, Ziyun Zeng, Yixiao Ge, Chen Li, Xintao Wang, and Ying Shan. 2023.
\newblock Making llama see and draw with seed tokenizer.
\newblock \emph{arXiv preprint arXiv:2310.01218}.

\bibitem[{Ge et~al.(2024)Ge, Zhao, Zhu, Ge, Yi, Song, Li, Ding, and Shan}]{ge2024seed}
Yuying Ge, Sijie Zhao, Jinguo Zhu, Yixiao Ge, Kun Yi, Lin Song, Chen Li, Xiaohan Ding, and Ying Shan. 2024.
\newblock Seed-x: Multimodal models with unified multi-granularity comprehension and generation.
\newblock \emph{arXiv preprint arXiv:2404.14396}.

\bibitem[{Ghosh et~al.(2023)Ghosh, Hajishirzi, and Schmidt}]{ghosh2023geneval}
Dhruba Ghosh, Hannaneh Hajishirzi, and Ludwig Schmidt. 2023.
\newblock Geneval: An object-focused framework for evaluating text-to-image alignment.
\newblock \emph{Advances in Neural Information Processing Systems}, 36:52132--52152.

\bibitem[{Guo et~al.(2025{\natexlab{a}})Guo, Yang, Zhang, Song, Zhang, Xu, Zhu, Ma, Wang, Bi et~al.}]{guo2025deepseek}
Daya Guo, Dejian Yang, Haowei Zhang, Junxiao Song, Ruoyu Zhang, Runxin Xu, Qihao Zhu, Shirong Ma, Peiyi Wang, Xiao Bi, and 1 others. 2025{\natexlab{a}}.
\newblock Deepseek-r1: Incentivizing reasoning capability in llms via reinforcement learning.
\newblock \emph{arXiv preprint arXiv:2501.12948}.

\bibitem[{Guo et~al.(2025{\natexlab{b}})Guo, Zhang, Tong, Zhao, Gao, Li, and Heng}]{guo2025generateimagescotlets}
Ziyu Guo, Renrui Zhang, Chengzhuo Tong, Zhizheng Zhao, Peng Gao, Hongsheng Li, and Pheng-Ann Heng. 2025{\natexlab{b}}.
\newblock \href {https://arxiv.org/abs/2501.13926} {Can we generate images with cot? let's verify and reinforce image generation step by step}.
\newblock \emph{Preprint}, arXiv:2501.13926.

\bibitem[{Han et~al.(2024)Han, Liu, Jiang, Yan, Zhang, Yuan, Peng, and Liu}]{han2024infinity}
Jian Han, Jinlai Liu, Yi~Jiang, Bin Yan, Yuqi Zhang, Zehuan Yuan, Bingyue Peng, and Xiaobing Liu. 2024.
\newblock Infinity: Scaling bitwise autoregressive modeling for high-resolution image synthesis.
\newblock \emph{arXiv preprint arXiv:2412.04431}.

\bibitem[{Hu et~al.(2024)Hu, Wang, Fang, Fu, Cheng, and Yu}]{hu2024ella}
Xiwei Hu, Rui Wang, Yixiao Fang, Bin Fu, Pei Cheng, and Gang Yu. 2024.
\newblock Ella: Equip diffusion models with llm for enhanced semantic alignment.
\newblock \emph{arXiv preprint arXiv:2403.05135}.

\bibitem[{Huang et~al.(2023)Huang, Sun, Xie, Li, and Liu}]{huang2023t2i}
Kaiyi Huang, Kaiyue Sun, Enze Xie, Zhenguo Li, and Xihui Liu. 2023.
\newblock T2i-compbench: A comprehensive benchmark for open-world compositional text-to-image generation.
\newblock \emph{Advances in Neural Information Processing Systems}, 36:78723--78747.

\bibitem[{Jiang et~al.(2025)Jiang, Guo, Zhang, Zong, Li, Zhuo, Yan, Heng, and Li}]{jiang2025t2i}
Dongzhi Jiang, Ziyu Guo, Renrui Zhang, Zhuofan Zong, Hao Li, Le~Zhuo, Shilin Yan, Pheng-Ann Heng, and Hongsheng Li. 2025.
\newblock T2i-r1: Reinforcing image generation with collaborative semantic-level and token-level cot.
\newblock \emph{arXiv preprint arXiv:2505.00703}.

\bibitem[{Labs(2024)}]{flux2024}
Black~Forest Labs. 2024.
\newblock Flux.
\newblock \url{https://github.com/black-forest-labs/flux}.

\bibitem[{Li et~al.(2024)Li, Zhang, Guo, Zhang, Li, Zhang, Zhang, Zhang, Li, Liu et~al.}]{li2024llava}
Bo~Li, Yuanhan Zhang, Dong Guo, Renrui Zhang, Feng Li, Hao Zhang, Kaichen Zhang, Peiyuan Zhang, Yanwei Li, Ziwei Liu, and 1 others. 2024.
\newblock Llava-onevision: Easy visual task transfer.
\newblock \emph{arXiv preprint arXiv:2408.03326}.

\bibitem[{Li et~al.(2023)Li, Pan, Ge, Gao, Ji, Zhang, Chua, Tang, Zhang, and Zhuang}]{li2023fine}
Juncheng Li, Kaihang Pan, Zhiqi Ge, Minghe Gao, Wei Ji, Wenqiao Zhang, Tat-Seng Chua, Siliang Tang, Hanwang Zhang, and Yueting Zhuang. 2023.
\newblock Fine-tuning multimodal llms to follow zero-shot demonstrative instructions.
\newblock \emph{arXiv preprint arXiv:2308.04152}.

\bibitem[{Li et~al.(2022)Li, Li, Xiong, and Hoi}]{li2022blip}
Junnan Li, Dongxu Li, Caiming Xiong, and Steven Hoi. 2022.
\newblock Blip: Bootstrapping language-image pre-training for unified vision-language understanding and generation.
\newblock In \emph{International conference on machine learning}, pages 12888--12900. PMLR.

\bibitem[{Lin et~al.(2025)Lin, Jia, Hu, Pan, Yue, Zhao, Chen, Wu, and Zhang}]{lin2025reasoning}
Wang Lin, Liyu Jia, Wentao Hu, Kaihang Pan, Zhongqi Yue, Wei Zhao, Jingyuan Chen, Fei Wu, and Hanwang Zhang. 2025.
\newblock Reasoning physical video generation with diffusion timestep tokens via reinforcement learning.
\newblock \emph{arXiv preprint arXiv:2504.15932}.

\bibitem[{Liu et~al.(2024{\natexlab{a}})Liu, Yan, Zaharia, and Abbeel}]{liu2024world}
Hao Liu, Wilson Yan, Matei Zaharia, and Pieter Abbeel. 2024{\natexlab{a}}.
\newblock World model on million-length video and language with blockwise ringattention.
\newblock \emph{arXiv preprint arXiv:2402.08268}.

\bibitem[{Liu et~al.(2024{\natexlab{b}})Liu, Li, Li, and Lee}]{liu2024improved}
Haotian Liu, Chunyuan Li, Yuheng Li, and Yong~Jae Lee. 2024{\natexlab{b}}.
\newblock Improved baselines with visual instruction tuning.
\newblock In \emph{Proceedings of the IEEE/CVF Conference on Computer Vision and Pattern Recognition}, pages 26296--26306.

\bibitem[{Liu et~al.(2023)Liu, Li, Wu, and Lee}]{liu2023visual}
Haotian Liu, Chunyuan Li, Qingyang Wu, and Yong~Jae Lee. 2023.
\newblock Visual instruction tuning.
\newblock \emph{Advances in neural information processing systems}, 36:34892--34916.

\bibitem[{Ma et~al.(2024)Ma, Liu, Chen, Liu, Wu, Wu, Pan, Xie, Zhang, yu, Zhao, Wang, Liu, and Ruan}]{ma2024janusflow}
Yiyang Ma, Xingchao Liu, Xiaokang Chen, Wen Liu, Chengyue Wu, Zhiyu Wu, Zizheng Pan, Zhenda Xie, Haowei Zhang, Xingkai yu, Liang Zhao, Yisong Wang, Jiaying Liu, and Chong Ruan. 2024.
\newblock Janusflow: Harmonizing autoregression and rectified flow for unified multimodal understanding and generation.

\bibitem[{OpenAI(2023)}]{openai2023chatgpt}
OpenAI. 2023.
\newblock Chatgpt.
\newblock \url{https://chat.openai.com}.

\bibitem[{OpenAI(2024)}]{gpt40}
OpenAI. 2024.
\newblock Introducing 4o image generation.
\newblock \url{https://openai.com/index/introducing-4o-image-generation/}.

\bibitem[{Pan et~al.(2024{\natexlab{a}})Pan, Fan, Li, Yu, Fei, Tang, Hong, Zhang, and Sun}]{pan2024towards}
Kaihang Pan, Zhaoyu Fan, Juncheng Li, Qifan Yu, Hao Fei, Siliang Tang, Richang Hong, Hanwang Zhang, and Qianru Sun. 2024{\natexlab{a}}.
\newblock Towards unified multimodal editing with enhanced knowledge collaboration.
\newblock \emph{Advances in Neural Information Processing Systems}, 37:110290--110314.

\bibitem[{Pan et~al.(2023)Pan, Li, Song, Lin, Liu, and Tang}]{pan2023self}
Kaihang Pan, Juncheng Li, Hongye Song, Jun Lin, Xiaozhong Liu, and Siliang Tang. 2023.
\newblock Self-supervised meta-prompt learning with meta-gradient regularization for few-shot generalization.
\newblock \emph{arXiv preprint arXiv:2303.12314}.

\bibitem[{Pan et~al.(2025{\natexlab{a}})Pan, Lin, Yue, Ao, Jia, Zhao, Li, Tang, and Zhang}]{pan2025generative}
Kaihang Pan, Wang Lin, Zhongqi Yue, Tenglong Ao, Liyu Jia, Wei Zhao, Juncheng Li, Siliang Tang, and Hanwang Zhang. 2025{\natexlab{a}}.
\newblock Generative multimodal pretraining with discrete diffusion timestep tokens.
\newblock \emph{arXiv preprint arXiv:2504.14666}.

\bibitem[{Pan et~al.(2024{\natexlab{b}})Pan, Tang, Li, Fan, Chow, Yan, Chua, Zhuang, and Zhang}]{pan2024auto}
Kaihang Pan, Siliang Tang, Juncheng Li, Zhaoyu Fan, Wei Chow, Shuicheng Yan, Tat-Seng Chua, Yueting Zhuang, and Hanwang Zhang. 2024{\natexlab{b}}.
\newblock Auto-encoding morph-tokens for multimodal llm.
\newblock \emph{arXiv preprint arXiv:2405.01926}.

\bibitem[{Pan et~al.(2025{\natexlab{b}})Pan, Wu, Bu, Shen, Li, Wang, Li, Tang, Xiao, Wu, Zhao, and Zhuang}]{pan2025unlocking}
Kaihang Pan, Yang Wu, Wendong Bu, Kai Shen, Juncheng Li, Yingting Wang, Yunfei Li, Siliang Tang, Jun Xiao, Fei Wu, Hang Zhao, and Yueting Zhuang. 2025{\natexlab{b}}.
\newblock Unlocking aha moments via reinforcement learning: Advancing collaborative visual comprehension and generation.
\newblock \emph{arXiv preprint arXiv:2506.01480}.

\bibitem[{Qiu et~al.(2024)Qiu, Gao, Qian, Pan, Yu, Li, Wang, Tang, Zhuang, and Chua}]{qiu2024step}
Haiyi Qiu, Minghe Gao, Long Qian, Kaihang Pan, Qifan Yu, Juncheng Li, Wenjie Wang, Siliang Tang, Yueting Zhuang, and Tat-Seng Chua. 2024.
\newblock Step: Enhancing video-llms' compositional reasoning by spatio-temporal graph-guided self-training.
\newblock \emph{arXiv preprint arXiv:2412.00161}.

\bibitem[{Shao et~al.(2024)Shao, Wang, Zhu, Xu, Song, Bi, Zhang, Zhang, Li, Wu et~al.}]{shao2024deepseekmath}
Zhihong Shao, Peiyi Wang, Qihao Zhu, Runxin Xu, Junxiao Song, Xiao Bi, Haowei Zhang, Mingchuan Zhang, YK~Li, Y~Wu, and 1 others. 2024.
\newblock Deepseekmath: Pushing the limits of mathematical reasoning in open language models.
\newblock \emph{arXiv preprint arXiv:2402.03300}.

\bibitem[{Sun et~al.(2024)Sun, Jiang, Chen, Zhang, Peng, Luo, and Yuan}]{sun2024autoregressive}
Peize Sun, Yi~Jiang, Shoufa Chen, Shilong Zhang, Bingyue Peng, Ping Luo, and Zehuan Yuan. 2024.
\newblock Autoregressive model beats diffusion: Llama for scalable image generation.
\newblock \emph{arXiv preprint arXiv:2406.06525}.

\bibitem[{Vice et~al.(2025)Vice, Akhtar, Hartley, and Mian}]{vice2025exploring}
Jordan Vice, Naveed Akhtar, Richard Hartley, and Ajmal Mian. 2025.
\newblock Exploring bias in over 100 text-to-image generative models.
\newblock \emph{arXiv preprint arXiv:2503.08012}.

\bibitem[{Wang et~al.(2024)Wang, Zhang, Luo, Sun, Cui, Wang, Zhang, Wang, Li, Yu et~al.}]{wang2024emu3}
Xinlong Wang, Xiaosong Zhang, Zhengxiong Luo, Quan Sun, Yufeng Cui, Jinsheng Wang, Fan Zhang, Yueze Wang, Zhen Li, Qiying Yu, and 1 others. 2024.
\newblock Emu3: Next-token prediction is all you need.
\newblock \emph{arXiv preprint arXiv:2409.18869}.

\bibitem[{Wu et~al.(2024)Wu, Zhang, Chen, Tang, Li, Fang, Zhu, Xie, Yin, Yi et~al.}]{wu2024vila}
Yecheng Wu, Zhuoyang Zhang, Junyu Chen, Haotian Tang, Dacheng Li, Yunhao Fang, Ligeng Zhu, Enze Xie, Hongxu Yin, Li~Yi, and 1 others. 2024.
\newblock Vila-u: a unified foundation model integrating visual understanding and generation.
\newblock \emph{arXiv preprint arXiv:2409.04429}.

\bibitem[{Xie et~al.(2025)Xie, Chen, Zhao, Yu, Zhu, Wu, Lin, Zhang, Li, Chen et~al.}]{xie2025sana}
Enze Xie, Junsong Chen, Yuyang Zhao, Jincheng Yu, Ligeng Zhu, Chengyue Wu, Yujun Lin, Zhekai Zhang, Muyang Li, Junyu Chen, and 1 others. 2025.
\newblock Sana 1.5: Efficient scaling of training-time and inference-time compute in linear diffusion transformer.
\newblock \emph{arXiv preprint arXiv:2501.18427}.

\bibitem[{Xie et~al.(2024)Xie, Mao, Bai, Zhang, Wang, Lin, Gu, Chen, Yang, and Shou}]{xie2024show}
Jinheng Xie, Weijia Mao, Zechen Bai, David~Junhao Zhang, Weihao Wang, Kevin~Qinghong Lin, Yuchao Gu, Zhijie Chen, Zhenheng Yang, and Mike~Zheng Shou. 2024.
\newblock Show-o: One single transformer to unify multimodal understanding and generation.
\newblock \emph{arXiv preprint arXiv:2408.12528}.

\bibitem[{Yin et~al.(2024)Yin, Zhao, Zhang, Lin, Wang, Tao, Wan, Zhang, Yin, and Zhang}]{yin2024sea}
Yuanyang Yin, Yaqi Zhao, Yajie Zhang, Ke~Lin, Jiahao Wang, Xin Tao, Pengfei Wan, Di~Zhang, Baoqun Yin, and Wentao Zhang. 2024.
\newblock Sea: Supervised embedding alignment for token-level visual-textual integration in mllms.
\newblock \emph{arXiv preprint arXiv:2408.11813}.

\bibitem[{Yu et~al.(2024)Yu, Chow, Yue, Pan, Wu, Wan, Li, Tang, Zhang, and Zhuang}]{yu2024anyedit}
Qifan Yu, Wei Chow, Zhongqi Yue, Kaihang Pan, Yang Wu, Xiaoyang Wan, Juncheng Li, Siliang Tang, Hanwang Zhang, and Yueting Zhuang. 2024.
\newblock Anyedit: Mastering unified high-quality image editing for any idea.
\newblock \emph{arXiv preprint arXiv:2411.15738}.

\bibitem[{Zhao et~al.(2024{\natexlab{a}})Zhao, Ma, Chen, Si, Wu, An, Yu, Zhang, Li, and Chang}]{zhao2024ultraedit}
Haozhe Zhao, Xiaojian~Shawn Ma, Liang Chen, Shuzheng Si, Rujie Wu, Kaikai An, Peiyu Yu, Minjia Zhang, Qing Li, and Baobao Chang. 2024{\natexlab{a}}.
\newblock Ultraedit: Instruction-based fine-grained image editing at scale.
\newblock \emph{Advances in Neural Information Processing Systems}, 37:3058--3093.

\bibitem[{Zhao et~al.(2024{\natexlab{b}})Zhao, Yin, Li, Lin, Huang, Chen, Chen, Yin, Zhou, and Zhang}]{zhao2024beyond}
Yaqi Zhao, Yuanyang Yin, Lin Li, Mingan Lin, Victor Shea-Jay Huang, Siwei Chen, Weipeng Chen, Baoqun Yin, Zenan Zhou, and Wentao Zhang. 2024{\natexlab{b}}.
\newblock Beyond sight: Towards cognitive alignment in lvlm via enriched visual knowledge.
\newblock \emph{arXiv preprint arXiv:2411.16824}.

\bibitem[{Zhuang et~al.(2025)Zhuang, Xie, Deng, Yang, Liang, Ru, Yin, and Zou}]{zhuang2025vargpt}
Xianwei Zhuang, Yuxin Xie, Yufan Deng, Dongchao Yang, Liming Liang, Jinghan Ru, Yuguo Yin, and Yuexian Zou. 2025.
\newblock Vargpt-v1. 1: Improve visual autoregressive large unified model via iterative instruction tuning and reinforcement learning.
\newblock \emph{arXiv preprint arXiv:2504.02949}.

\end{thebibliography}

\appendix
\newpage
\section*{Appendix}
\section{More Details on PairComp}
\label{app1}

\definecolor{mycolor}{rgb}{0.122, 0.435, 0.698}
\newtcolorbox{mybox}{colframe =mycolor}

Each test case in Paircomp contains two similar prompts with subtle differences. 
The two prompts exhibit word-level differences that lead to noticeable distinctions in six types of fine-grained semantic aspects: (1) Overall appearance difference; (2) Color difference; (3) Counting difference; (4) Position difference; (5) Style \& Tone difference; (6) Text difference. Next, we will provide a detailed explanation of these six types.
\begin{itemize}

\item \textbf{Color:} highlighting differences in the color of specific items in two images. For example, an umbrella in one picture is purple while in another picture it is green. 

\item \textbf{Position:} Differences in the relative positioning of specific items in two images. For example, in one picture object \texttt{[A]} is to the left of object \texttt{[B]} while in another picture \texttt{[A]} is to the right of \texttt{[B]}. 

\item \textbf{Text:} Differences in the textual content on an item in two images. For example, the departure time on a ticket in one picture is "20:00" while the departure time on a ticket in another picture is "21:00". 

\item \textbf{Style \& Tone:} The differences can be categorized into two types: (1) Differences in the overall style of two images. For example, one picture is in an oil painting style while another picture is in an ink wash painting style. (2) Differences in the overall atmosphere (weather, season, etc.) in two images. For example, the scene depicted in one picture is on a sunny day while the scene depicted in another picture is on a foggy day. 

\item \textbf{Counting:} Differences in the quantity of specific items in two images. For example, there are 3 eggs in one picture while there are only 2 eggs in another picture. 

\item \textbf{Overall-appearance:} Differences in the overall appearance of items in two images, including but not limited to the previously mentioned item such as color, as well as previously unmentioned decorations or style differences of objects. For example, a cat in one picture is wearing a bow tie while a cat in another picture is wearing a bell.
\end{itemize}

\section{More Details on FocusDiff-Data}
\label{app2}
\begin{figure}[t]
\includegraphics[width=\linewidth]{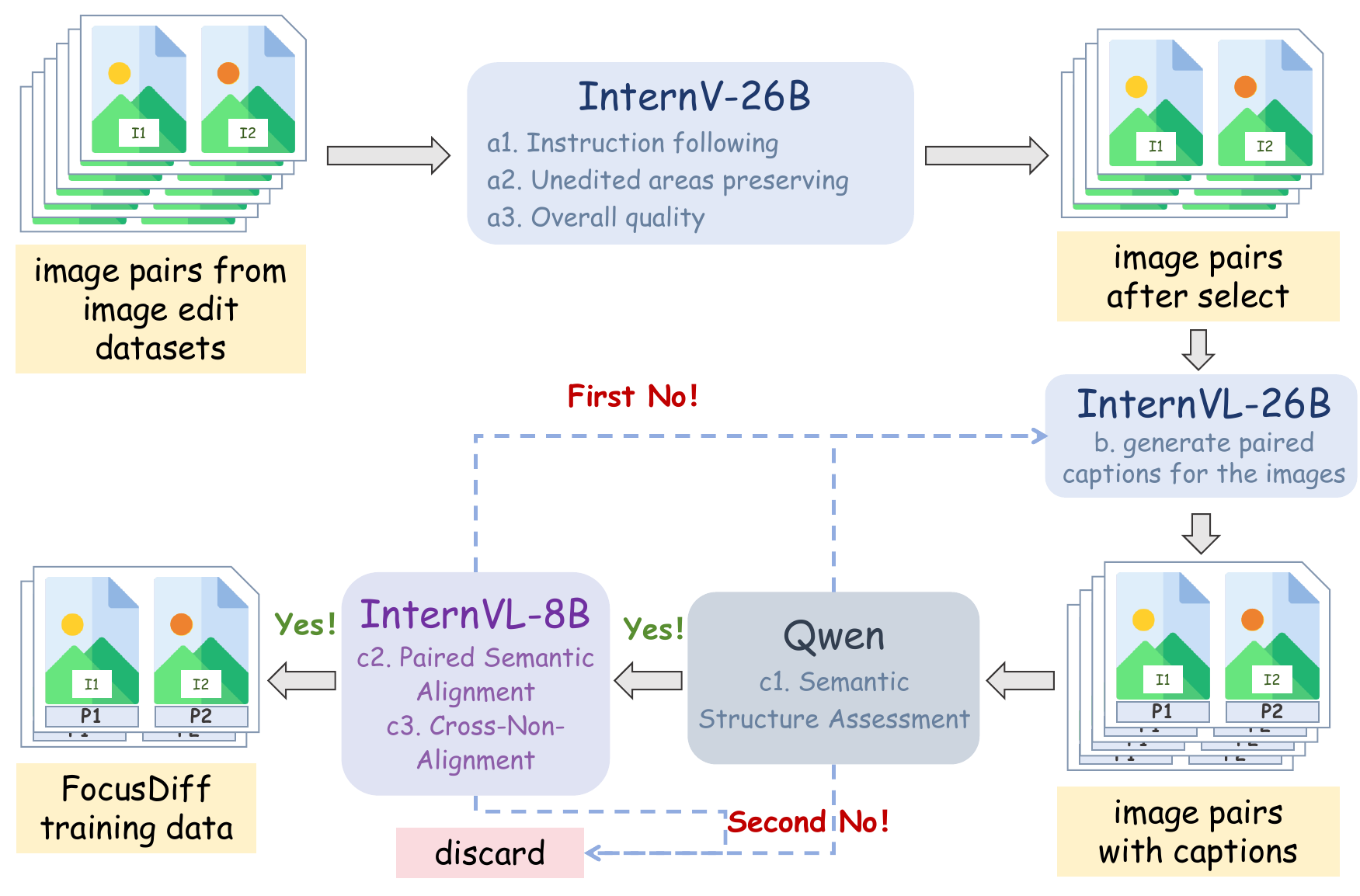}
\vspace{-1.5em}
\centering\caption{The pipeline for constructing \texttt{FocusDiff-Data}}
\label{fig:intern}
\vspace{-1em}
\end{figure}

In this section, we give more details on how to construct \texttt{FocusDiff-Data} from the image editing dataset~\cite{zhao2024ultraedit, yu2024anyedit}, with the pipeline shown in \ref{fig:intern}.
In the first step, considering the potential poor quality of the image editing dataset, we conduct data cleaning on the raw data to retain only high-quality samples.
Using the InternVL2.5-26B model, providing it with the before-after-editing images and the editing instruction, we evaluate three key aspects with the following prompts: \textit{\textbf{(1)} whether the edited image follows the editing instructions}; \textit{\textbf{(2)} whether the non-edited areas of the edited image remain consistent with the original image}; and \textit{\textbf{(3)} whether the overall quality and natural appearance of the edited image are acceptable}. We filter out any pairs that fail to meet these criteria.

Subsequently, we input the pair of before-and-after images along with the editing instructions into InternVL2.5-26B~\cite{chen2024internvl}. 
We prompt it to generate a pair of captions for the images that share a similar stylistic structure but differ only in individual words, thereby highlighting the differences between the images. The task prompt is formatted as:
\begin{mybox}
The user will provide an original image and an edited image based on specific editing instructions. Your task is to write a description for each of these two images. The descriptions must adhere to the following guidelines.\newline
\textbf{Identical Structure:} Both descriptions should follow the exact same structural format. Ensure that verbs, adjectives, and other parts of speech align in number and position between the two sentences.\newline
\textbf{Minimal Differences:} Only one to three words should be altered between the original and edited image descriptions to emphasize the changes made.\newline
\textbf{Direct Comparison:} The paired sentences should correspond directly, allowing for a clear comparison between the original and edited images.
\end{mybox}

After generating the captions $(\mathcal{P}_1,\mathcal{P}_2)$ for the images $(\mathcal{I}_1,\mathcal{I}_2)$, we conduct a post-verification operation with three conditions: 
\textbf{(1)} Using the Qwen model~\cite{qwen}, we assess whether $\mathcal{P}_1$ and $\mathcal{P}_2$ exhibit similar semantic structures; 
\textbf{(2)} Using the InternVL-8B model~\cite{chen2024internvl}, we verify whether $\mathcal{P}_1$ and $\mathcal{I}_1$, as well as $\mathcal{P}_2$ and $\mathcal{I}_2$, were semantically aligned. 
\textbf{(3)} We further leverage InternVL-8B to ensure that  $\mathcal{P}_1$ and $\mathcal{I}_2$, as well as  $\mathcal{P}_2$ and $\mathcal{I}_1$, are not semantically aligned.
If all of three conditions are satisfied, the sample is deemed valid and included in our training dataset. Otherwise, we request the InternVL2.5-26B model to regenerate captions for the two images and conduct the post-verification again. If the post-verification still fails, the image pair is then discarded.
Finally, we retained approximately $200,000$ high-quality data pairs.

\section{Implementation Details}
\label{app3}
\paragraph{Supervised Fine-Tuning.} We first leverage FocusDiff-Data to conduct autoregressive text-to-image supervised fine-tuning on Janus-Pro. The objective function is $p(y) = \frac{1}{S} \sum_{i=1}^{S} \log P_\theta(y_i | y_{< i},\mathcal{T})$, where $y$ is the visual token of an image with $S$ as the sequence length, $\mathcal{T}$ is the text condition. The detailed hyperparameters for training are shown in Table~\ref{tab:hyper}.

\paragraph{Reward Calculation.}
The overall design philosophy of our reward model is to leverage QA-based visual comprehension~\cite{chen2024internvl, li2023fine, pan2024towards, li2024llava, qiu2024step, liu2023visual, liu2024improved,chen2024improving} models, which will return a consistency score $\texttt{R}^{QA}(\cdot)\in [0,1]$ for each text-image pair.
We leverage  InternVL2.5-26B~\cite{chen2024internvl} as the reward model to provide appropriate incentives, which will return a consistency score $\mathtt{R}_{\mathcal{I}}\in [0,1]$ for each text-image pair.
Specifically, for short prompts, we directly the MLLM with the question ``\texttt{Does this image match the description? Please directly respond with yes or no}''.  We record the probability of the model responding with ``Yes'' as $P_{yes}$ and ``No'' as $P_{no}$, with the reward score calculated as $S(\mathcal{I}, \mathcal{P}) = P_{yes} / (P_{yes} + P_{no})$.
For long prompts, inspired by \citet{}, we first decompose the prompt to semantic tuples (\textit{e.g.}, attribute, and spatial relation) and then generate yes-or-no questions (\textit{e.g.}, ``Is the cdog red?''). 
The MLLMs are asked to perform a VQA task for the prompt and generated image, returning a score of 0 to 1 for each question in the same way.
The reward is obtained by averaging the evaluation of
the MLLMs on multiple questions for a prompt.

\paragraph{Reinforcement Learning.}

\begin{table}[t]
    \centering
    \caption{\label{tab:hyper}The detailed training hyper-parameters of supervised fine-tuning and reinforcement learning.}
    \vspace{-1em}
    \begin{adjustbox}{width=0.5\textwidth}
    \begin{tabular}{lcc}
    \toprule
    \textbf{Hyper-parameters} & \textbf{Fine-Tuning} & \textbf{Reinforcement Learning} \\ \hline
    Optimizer & AdamW & AdamW  \\
    Optimizer param. & \multicolumn{2}{c}{$\beta_1=0.9,\beta_2=0.95,\epsilon=1\mathrm{e}{-6}$}  \\
    Peak LR & 2.0e-5 & 1.0e-5  \\
    Convert LR & - & 2.0e-6 \\
    Convert step & - & 300 \\
    Min LR & 2.0e-7 & 2.0e-7 \\
    LR scheduler & Cosine & Linear+Cosine  \\
    Batch size & 256 & 128 \\
    Training Steps & 5K & 2.2K  \\
    Warmup Steps & 100 & 100 \\
    Weight decay   & 0.05 & 0.05 \\
    Gradient clipping    & 1.0 & 1.0 \\
    Numerical precision    & bfloat16 & bfloat16  \\
    Resource Usage    & 8 NVIDIA A800 & 16 NVIDIA A800 \\
    \bottomrule
    \end{tabular}
    \end{adjustbox}
    \vspace{-1em}

\end{table}
Our proposed Pair-GRPO is an improved version of GRPO, with training prompts sourced from FocusDiff-Data. 
We set the $G=7$, first expanding the group size from $7$ to $14$. 
There is a probability $p$ that the group size may further increase to $18$, as we introduce ground-truth images corresponding to prompt pairs from FocusDiff-Data and pair them with the prompts. The probability $p$ is dynamic, incorporating the concept of curriculum learning~\cite{bengio2009curriculum,pan2023self}, decreasing from 1.0 at the start of training to 0.0 by the end.

During RL training, we used the fine-tuned Janus-Pro as the backbone model and set the batch size to 128, meaning that each optimization iteration includes 128 different prompts. All parameters are tunable. We totally conduct 2200 iterations of post-training optimization. 
We find that the learning rate is crucial: a learning rate that is too small results in insignificant performance gains, while a learning rate that is too large leads to unstable training. 
To address this, we design a combined Linear + Cosine learning rate scheduler. 
The learning rate quickly drops linearly from a peak value to a lower ``convert learning rate'' at a ``convert step'', and then gradually decreases along a cosine curve.
However, we still encounter some instability during training, indicated by a downward trend in the reward curve. 
To address this, we adopt the following measures: 
(1) When the reward curve dropped sharply, we reduce the learning rate to half or two-thirds of its current value and resume the training; 
(2) When the reward curve declined gradually, it suggests that the KL divergence constraint with a less capable reference model is  limiting the model improvement. Thus we update the reference model to the current model and then resume the training.
The detailed hyperparameters for training are shown in Table~\ref{tab:hyper}.

\paragraph{Inference.} During inference, we follow the inference setup of Janus-Pro, setting $topk = 4096$ for visual token sampling. 
Besides, we use classifier-free guidance on the logits for autoregressive sampling in a manner similar to \cite{pan2025generative, wang2024emu3, liu2024world}. We set the guidance scale to 5.0 or 6.0.

\section{Evaluation Details}
\label{app4}
\paragraph{Baseline.}
we compare Janus-Pro-R1 with both diffusion-based and AR-based methods. 
The diffusion-based baselines include PixArt-alpha~\cite{chen2023pixartalphafasttrainingdiffusion}, DALL-E3~\cite{betker2023improving}, SD3~\cite{esser2024scaling}, FLUX.1-dev~\cite{flux2024}, Sana-1.5~\cite{xie2025sana}, and Janus-Flow~\cite{ma2024janusflow}. 
The AR-based baselines include LLamaGen~\cite{sun2024autoregressive}, VILA-U~\cite{wu2024vila}, Show-o~\cite{xie2024show}, SEED-X~\cite{ge2024seed}, Emu-3~\cite{wang2024emu3}, DDT-LLaMA~\cite{pan2025generative}, VARGPTv1.1~\cite{zhuang2025vargpt}, Infinity~\cite{han2024infinity}, Janus-Pro~\cite{chen2025janus}, BLIP3-o~\cite{chen2025blip3}, GPT-4o~\cite{gpt40}, Show-o+PARM~\cite{guo2025generateimagescotlets}, T2I-R1~\cite{jiang2025t2i}, and Janus-Pro-R1~\cite{pan2025unlocking}.
It is worth noting that, Show-o+PARM, T2I-R1 and Janus-Pro-R1 attempt to enhance the text-to-image generation capabilities of AR-based MLLMs through reinforcement learning. Furthermore, among these baselines, we only report the performance of open-source models for the PairComp benchmark.

\paragraph{Benchmarks.}
In PairComp, we leverage InternVL2.5-26B as the evaluation model with the prompt: ``Does this image match the description? Please directly respond with yes or no.''
We record the probability of the model responding with ``yes'' (denoted $P_{yes}$) and with ``no'' (denoted $P_{no}$), with the semantic consistency score calculated as $S(\mathcal{I}, \mathcal{T}) = P_{yes} / (P_{yes} + P_{no})$.
For each prompt, we require a text-to-image model to generate two images. Therefore, for a pair of similar prompts $(\mathcal{T}^{1}_i,\mathcal{T}^{2}_i)$, we obtain four generated images $(\mathcal{I}^{1,1}_i,\mathcal{T}^{1,2}_i, \mathcal{I}^{2,1}_i, \mathcal{I}^{2,2}_i)$.
We then compute the semantic consistency scores for each image with respect to its corresponding prompt: $s^{1,1}_i=S(\mathcal{I}^{1,1}_i, \mathcal{T}^{1}_i)$, $s^{1,2}_i=S(\mathcal{I}^{1,2}_i, \mathcal{T}^{1}_i)$, $s^{2,1}_i=S(\mathcal{I}^{2,1}_i, \mathcal{T}^{2}_i)$, $s^{2,2}_i=S(\mathcal{I}^{2,2}_i, \mathcal{T}^{2}_i)$.
The arithmetic mean score is calculated as:
$s_a = \frac{1}{4N} \sum_{i=1}^N( s_i^{1,1}+s_i^{1,2}+s_i^{2,1}+s_i^{2,2})$,
and the geometric mean score is calculated as: $s_g = \frac{1}{N} \sqrt[4]{ s_i^{1,1}\cdot s_i^{1,2}\cdot s_i^{2,1}\cdot s_i^{2,2}} $.
The score of the geometric (arithmetic) mean for ``Average'' is obtained by averaging the geometric (arithmetic) mean scores of the other six sub-tasks.

Furthermore, we also conduct zero-shot evaluation on 3 existing text-to-image benchmarks: GenEval~\cite{ghosh2023geneval}, T2I-CompBench~\cite{huang2023t2i}, and DPG-Bench~\cite{hu2024ella}. 
GenEval contains 6 different subtasks of varying difficulty requiring various compositional skills, including \texttt{single object} (SingObj), \texttt{single object} (TwoObj), \texttt{counting}, \texttt{colors}, \texttt{position}, \texttt{color binding} (ColorAttri). 
And we adopt the metric proposed by ~\cite{ghosh2023geneval} for evaluation.
Each subtask is scored independently, and the overall score is calculated as the average of all six subtask scores. 
The T2I-CompBench encompasses three subtasks following \citet{wang2024emu3}: \texttt{color}, \texttt{shape}, \texttt{texture}. Building on prior research, we employ the Blip-VQA score~\cite{li2022blip}  as the evaluation metric.
While for DPG-Bench, we follow the metrics proposed in \cite{hu2024ella} to conduct evaluation.

\end{document}